%% file: MAIN_COLING.tex
\documentclass[10pt, a4paper]{article}

\usepackage{lrec-coling2024} % this is the new style
\usepackage{inconsolata}

\usepackage{comment}
\usepackage{booktabs}
\usepackage{tabularx}
\usepackage{graphicx}
\usepackage{multirow}
\usepackage{colortbl}
\usepackage{enumitem}

\usepackage{indentfirst}
\usepackage{caption}
\usepackage{subcaption}
\usepackage{xurl}

\definecolor{SubtleBlue}{rgb}{0.9, 0.95, 1.0}
\definecolor{HeaderBlue}{rgb}{0.8, 0.9, 1.0}

\newcommand{\tabitem}{\textbullet~~}
\title{ChatGPT Role-play Dataset: Analysis of User Motives and Model Naturalness}

\name{Yufei Tao$^{\diamond,1}$, Ameeta Agrawal$^{\diamond,2}$, Judit Dombi$^{\ddagger,3}$, Tetyana Sydorenko$^{^\circ,4}$, Jung In Lee$^{^\circ,5}$} 

\address{$^\diamond$ Department of Computer Science, Portland State University, Portland, USA\\
$\ddagger$Department of English Linguistics, University of Pécs, Pécs, Hungary\\
$^\circ$Department of Applied Linguistics, Portland State University, Portland, USA\\
{\{\texttt{$^1$yutao, $^2$ameeta, $^4$tsydorenko, $^5$jungin\}@pdx.edu, $^3$dombi.judit@pte.hu}}}
\abstract{
Recent advances in interactive large language models like ChatGPT have revolutionized various domains; however, their behavior in natural and role-play conversation settings remains underexplored. In our study, we address this gap by deeply investigating how ChatGPT behaves during conversations in different settings by analyzing its interactions in both a normal way and a role-play setting. We introduce a novel dataset of broad range of human-AI conversations annotated with user motives and model naturalness to examine (i) how humans engage with the conversational AI model, and (ii) how natural are AI model responses.  Our study highlights the diversity of user motives when interacting with ChatGPT and variable AI naturalness, showing not only the nuanced dynamics of natural conversations between humans and AI, but also providing new avenues for improving the effectiveness of human-AI communication.
 \\ \newline  }

\begin{document}

\maketitleabstract

\input{1-intro}

\input{2-related}

\input{3-data}

\input{4-method}

\input{5-analysis}
\input{6-discussion}

\input{7-conclusion}

% \nocite{*}
\section{Bibliographical References}\label{sec:reference}

\bibliographystyle{lrec-coling2024-natbib}
\bibliography{lrec-coling2024-example, custom}

\end{document}

%% file: 1-intro.tex
\section{Introduction}

%(Outline: novelty, question, usefulness of the findings, overview of methodology, results)

Although conversational systems have been around for decades \cite{weizenbaum1966eliza}, in the last few years the natural language processing (NLP) capabilities have greatly improved, to the point where  interactive large language models (LLM), such as ChatGPT by OpenAI, are making headlines. % in the general population, industry, and academia. 
Studies  have focused on quantitative evaluation of ChatGPT on numerous NLP tasks \cite{bang2023multitask}, qualitative assessment \cite{thorp2023chatgpt}, or examining its role in various applications and domains \cite{shahriar2023let}. However, research with human-produced data studying human-AI communication is scarce, and systematically studying the behavior of these LLMs in interactional contexts is even more challenging.

In this study, we explore two questions: (i) how humans interact with a conversational AI model (ChatGPT), and (ii) whether the AI model can be conversational enough to  provide the specific benefit of human-like conversation. The former entails {\em user motives}, or in other words users' conversational intents, and is informed by prior research on how humans perceive interactions with machines \cite{Nass2000}. %lee2010trust \ts{we can delete Lee reference to save space if you like} 
The second question pertains to the {\em naturalness of the model's responses} and is informed by prior work on the rules of human conversation \cite{grice1975logic, grice1989studies}. People may have a variety of reasons to practice human-like conversations with a machine (e.g., students role-playing challenging conversations to learn and explore information \cite{roleplaying} or medical students practicing doctor-patient interactions \cite{eysenbach2023role}).

%(e.g., foreign language learners practicing conversational skills (Sydorenko et al, in press) or medical students practicing doctor-patient interactions \cite{eysenbach2023role}).

%The former entails {\em user motives} and tendency to give feedback to an AI system, whereas the latter pertains to the {\em model's naturalness}. These two go hand in hand as we hypothesize that humans would primarily want to converse with the AI model. 

\begin{figure}[t!]
\centering
\vspace{0.2cm}
\includegraphics[width=0.5\textwidth]{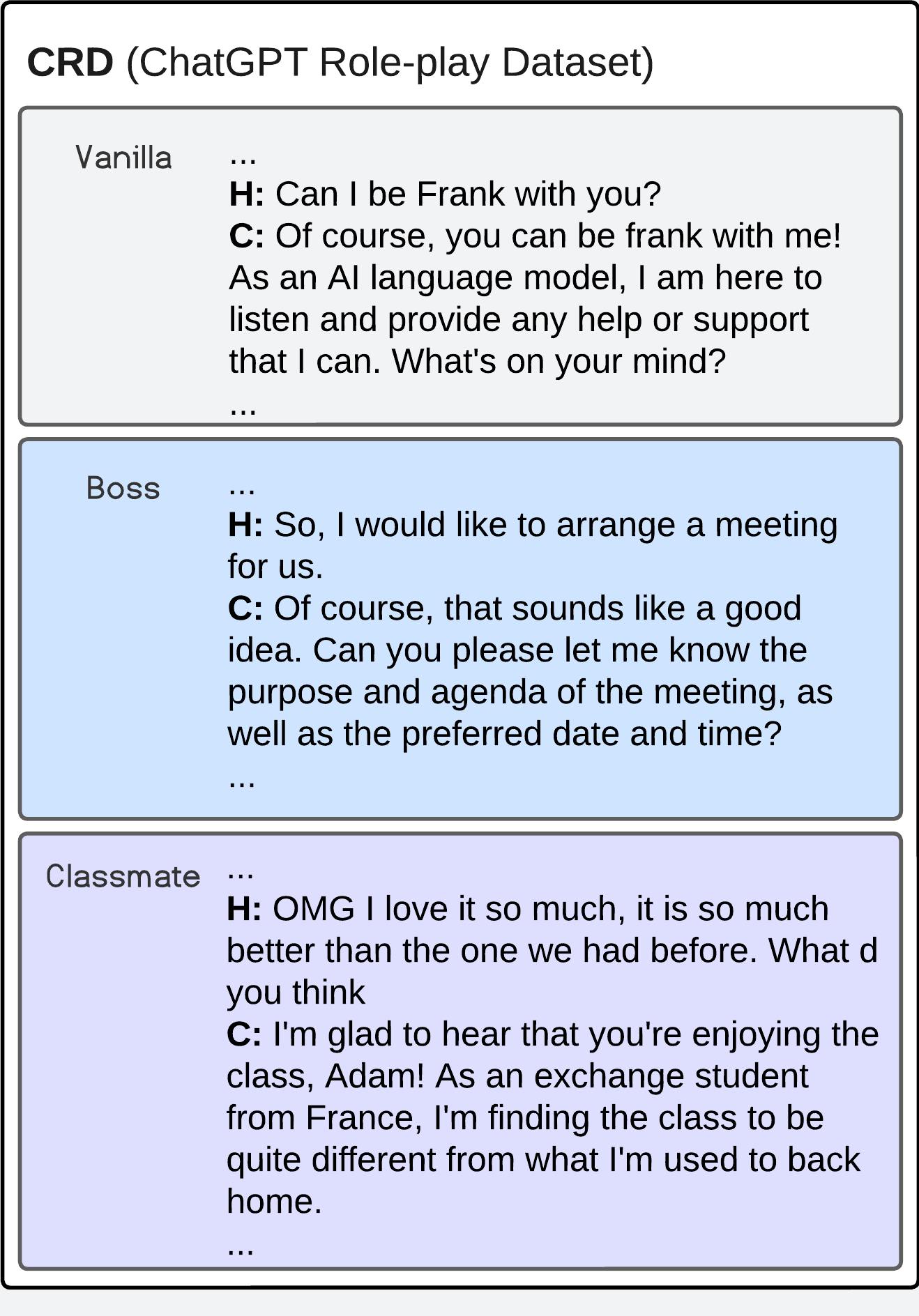}

\caption{Snippets of conversations from our dataset CRD, where \texttt{vanilla}, \texttt{boss}, and \texttt{classmate} denote the three subsets of CRD. `H' denotes the human utterance, whereas `C' indicates the response generated by  ChatGPT.} \label{fig:snippets}
%\vspace{-0.5cm}
\end{figure}
 
%Conversational models can be used for various purposes, one of which is second language learning \cite{Huang2022Chatbots}. Various second language learning theories (such as the Interaction Approach and the Sociocultural Theory) indicate that interactions in a second language between humans can effectively support language development \cite{Mitchell2022Second}. %It is especially beneficial when humans give feedback to language learners.

%novelty

%novelty: Interactive data from two different student/participant (?) populations human annotators annotated the data following Grice’s maxims and more
{We introduce a new conversational dataset -- {ChatGPT Role-play Dataset (CRD)\footnote{\url{https://github.com/PortNLP/ChatGPT_Role-play_Dataset}}}} -- consisting of three distinct subsets of conversations, drawing interactive data from participants that engaged with ChatGPT under regular and different {role-play} settings. {Given the  limited availability of resources for studying ChatGPT's role-play capabilities, focusing on and analyzing conversations under these settings is particularly important.} Some excerpts from these conversations are provided in Figure~\ref{fig:snippets}. We manually annotate the conversations in CRD for user motives and model naturalness, making it the first dataset of its kind, to our knowledge. %We also incorporate Grice's maxims of conversation during our analysis which allow  us to capture the nuances of interaction between humans and a conversational AI model. %This approach results in datasets that more accurately represent the dynamics of human-AI conversation.

%(Cai, and Mitrovic). =====

%question

%	We can rephrase number (2) to something else, but I think it’s important to include- that’s the part about bot’s naturalness. (Number 1 is about user motives and giving feedback). They kind of go hand in hand since the human does seem to want the bot to be conversational, given a high number of “convo” user motives. 

%describe potential benefits of the findings and applications: 

%What it means for future AI systems: 
%What it tells us about human communication with an AI model:
%Whether human-human communication compares to human-AI communication

%We did not explore ChatGPT’s ability to provide human-like feedback, although this can be the next logical step.]]

%== overview of methodology ==
We conduct a wide range of analyses, both statistical and those grounded in linguistic theories. Because our intention was to gather conversational data (rather than, for example, to examine how well ChatGPT can write essays), in our analyses we relied on Gricean pragmatics which looks at the nuances of intention behind human utterances as well as cooperation in communication. Besides, we also relied on partner modeling \cite{HortonConversational2005} and communication accommodation theory \cite{giles_coupland_coupland_1991} to understand how users conceptualize their AI communication partner and how they adjust their utterances to the context.

The key contributions of our research include:

 \begin{itemize}
     \item A novel dataset -- ChatGPT Role-play Dataset (CRD) -- comprising 85 unique conversations with ChatGPT, resulting in a total of 1742 utterances. We manually annotate every single utterance for user motives and model naturalness;

    \item An extensive analysis of human-AI conversations to understand communicative intentions.
    
\end{itemize}

%% file: 2-related.tex
\section{Related Work}

\noindent \textbf{\em ChatGPT}. \quad ChatGPT is a generative large language model enhanced with multi-turn dialogue capabilities from the GPT-3.5 and GPT-4 series in OpenAI’s family of GPT models \cite{radford2018improving, radford2019language, brown2020language, ouyang2022training}.

%following several earlier models such as GPT, GPT-2, GPT-3

%such as GPT \cite{radford2018improving}, GPT-2 \cite{radford2019language}, GPT-3 \cite{brown2020language}, and InstructGPT \cite{ouyang2022training}.

%%empirical analysis: Gozalo-Brizuela and Garrido-Merchan, 2023 ChatGPT is not all you need. A State of the Art Review of large Generative AI models
%education Guo et al., 2023), 
%medical writing (Biswas, 2023). 
%and a qualitative analysis using “red-teaming” (Zhuo et al., 2023). 

Previous efforts to study ChatGPT in different aspects range from empirical analysis in the realm of NLP tasks \cite{bang2023multitask, azaria2022chatgpt, qin2023chatgpt, wang2023robustness, lai2023chatgpt, tabone2023using} to interest in other areas of research such as regulation \cite{hacker2023regulating},  ethics \cite{zhuo2023exploring}, law \cite{choi2023chatgpt}, education \cite{susnjak2022chatgpt}, and medicine \cite{jeblick2022chatgpt}. Previous work also includes a meta-analysis on how ChatGPT is perceived and discussed \cite{leiter2023chatgpt}, a description of relevant research priorities \cite{van2023chatgpt}, and its cultural implications \cite{cao2023assessing}. 

%such as regulation \cite{hacker2023regulating, shen2023chatgpt},  ethics \cite{zhuo2023exploring}, law \cite{choi2023chatgpt}, education \cite{khalil2023will, m2022exploring, susnjak2022chatgpt}, and medicine \cite{jeblick2022chatgpt}. Prior work also includes a meta-analysis on how ChatGPT is perceived and discussed \cite{leiter2023chatgpt}, an outline of relevant research priorities \cite{van2023chatgpt}, and its cultural implications \cite{cao2023assessing}.

Several attempts have also been made to distinguish between human-written and ChatGPT-generated text \cite{mitrovic2023chatgpt, guo2023close, pegoraro2023chatgpt, islam2023distinguishing}. However, none of this research has examined true human-computer interacton in real time. %For instance, with an analysis primarily done using explainable AI, \citet{mitrovic2023chatgpt} observed that ChatGPT’s writing is polite, without specific details, with fancy and atypical vocabulary, impersonal, and typically does not express feelings. 

\medskip

\noindent \textbf{\em Communicative interactions}. \quad While earlier work on the pragmatics of human-computer interaction has been theoretical \cite{searle1984minds}, some studies have drawn upon experimental data \cite{Wolters2009ReducingWM, Georgila2010TheMC,Chai2014CollaborativeET, Fischer2016DesigningSpeech, FischerSituatedness2017}. 
%on human-robot interactions \cite{Chai2014CollaborativeET, Fischer2016DesigningSpeech, FischerSituatedness2017} and human-spoken dialog system interactions \cite{Wolters2009ReducingWM, Georgila2010TheMC}. 
Most recently, with the rapid advancement in interactive LLMs, there is strong interest in examining whether LLMs resemble humans in their language use \cite{cai2023does, chomsky2023false, mahowald2023dissociating, piantadosia2023modern,kasirzadeh2023conversation, Sydorenko2023}.

Human communication is an intention-directed practice, during which the interlocutors mutually recognize their intentions and goals, and make joint efforts to achieve them \cite{Clark1996UsingL}. In communication, interactants subconsciously and automatically cooperate \cite{grice1975logic, grice1989studies}. 
Understanding intentions and deriving indirect meanings, such as conversational implicatures, are thus central to human communication, yet a difficult task for LLMs \cite{ruis2022large, Kim2023IsTP, qiu_duan_cai_2023}. ChatGPT replicates many patterns of human language use, among others, it reuses recent sentence structures, reinterprets implausible sentences corrupted by noise, glosses over errors and draws reasonable inferences \cite{cai2023does}, {and to some extent, captures embedded meaning \cite{Marchetti2023Mind}. Moreover, efforts have been made to enhance ChatGPT's performance in role-playing conversation settings, underscoring its evolving ability to mimic human-like interactions more closely \cite{kong2023better, shanahan2023roleplay, lu2024large}. Despite these advancements, ChatGPT's responses still}
% or captures embedded meaning to some extent \cite{Marchetti2023Mind}. \yt{Serveral works have also been done on making ChatGPT better in role-playing conversation settings. \cite{kong2023better, shanahan2023role-play, lu2024large}}. On the other hand, its answers 
rarely appear as natural, as they are oftentimes rather long, hesitant or even obscure \cite{Brunet2023theory}; in linguistic terms ChatGPT does not observe the Gricean maxims of Quantity, Quality, Manner and Relevance.

Given the aforementioned findings that ChatGPT’s responses are often not entirely human-like,
our work examines the human intentions behind utterances directed to an AI model and the AI’s ability to uptake the human intent and to appropriately answer it approximating human natural responses. We are also curious to find out if humans provide feedback to the AI model on how they feel about the conversation.

%% file: 3-data.tex
\section{CRD: ChatGPT Role-play Dataset}

Here we describe our process of collecting and annotating several conversations with ChatGPT.

\subsection{Data Collection}
The data collection part involved participants who interacted with ChatGPT during March - April 2023\footnote{For this study, ChatGPT-3.5 March 13, 2023 and March 23, 2023 versions were used.}. Collectively, our dataset includes 57 participants, 85 unique conversations, and a total of 1742 utterances. Participants were largely comparable in age and prior experience using text chatbots, but showed notable variation in their first languages which included 15 different languages (Arabic, Bengali, Chinese, English, Gujarati, Hindi, Hungarian, Khmer, Magyar, Malay, Nepali, Persian, Sakha, Tamil, Telugu). 

%{(more details in Appendix~\ref{app:students})}

CRD consists of conversations conducted in two settings:

\begin{itemize}
\item \textbf{\texttt{vanilla}} \quad %Computer science   (N = 29)
Participants interacted with ChatGPT ``as is'' for 5-10 minutes. The participants were asked to insert indirect statements to challenge ChatGPT (e.g., jokes, sarcasm, metaphors). The goal of this scenario was to understand how humans engage with a novel conversational AI model and how well the model can understand and respond to various intents.

\item \textbf{(Role-play) \texttt{boss} and \texttt{classmate} } \quad %English language learners 
Participants interacted with ChatGPT for 5-10 minutes in two role-play settings: once where the model played the role of a boss and another time as a classmate, following the prompts developed by the authors\footnote{Prompt for \texttt{boss}: {\em Could we do a role-play where you are my boss and I ask you a question, and my boss's name is Lisa? In your responses, please don't say you are an AI model, OK?}

Prompt for \texttt{classmate}: {\em Can we do a role-play where you and I are classmates in an English language class? Your name is Florian, and you are an exchange student from France. Please make up all the facts about Florian. It is break time and we decided to strike a conversation. You and I are talking for the first time. In your responses, please don't say you are an AI model, OK? I will start the conversation. Hi, I am Adam. What’s your name?}
}.

%Because not all users approach ChatGPT with a uniform intent, we seek to examine ChatGPT's conversational capabilities, given the significance of conversations in language learning. 

The two role-plays are interesting for two reasons. In pragmatics, context is central, and we use two very different contexts to examine a wide range of social and linguistic behavior. There are three social variables based on which speakers may vary their strategies in interactions: social distance, power, and imposition \cite{brown1987politeness}.  Social distance means differing degrees of familiarity between interlocutors, power means the addressee’s position in society, and imposition is about the risk posed by the utterance.  

%So in the two scenarios we basically played with the social variables that may have an impact on human interactional behavior. 

In \texttt{boss} there is a request, which is a high imposition feature, and the power distance is uneven. These factors make the scenario face-threatening. In \texttt{classmate}, the power factor is even, and the degree of imposition is unspecified, which creates a much less face-threatening situation. In addition, the boss role provides a structured and predictable task, while the classmate role is more open-ended. 
% -- We could argue that the social distance in \texttt{classmate} is higher, as participants were to meet this new international student (role-played by ChatGPT), whereas they likely knew their boss, but even without this line of argumentation the two scenarios are meaningful to compare being ``high stakes" vs ``low-stakes" conversationally. % -- very much likely to be face threatening for the Boss and not quite likely in the Class scenario. 
As such, these two role-playing settings provide a diverse yet controlled environment for our analysis. %`boss' and `classmate' represent two distinct conversational dynamics. 

\end{itemize}

\begin{figure}[t!]
\centering
\includegraphics[width=0.55\textwidth]{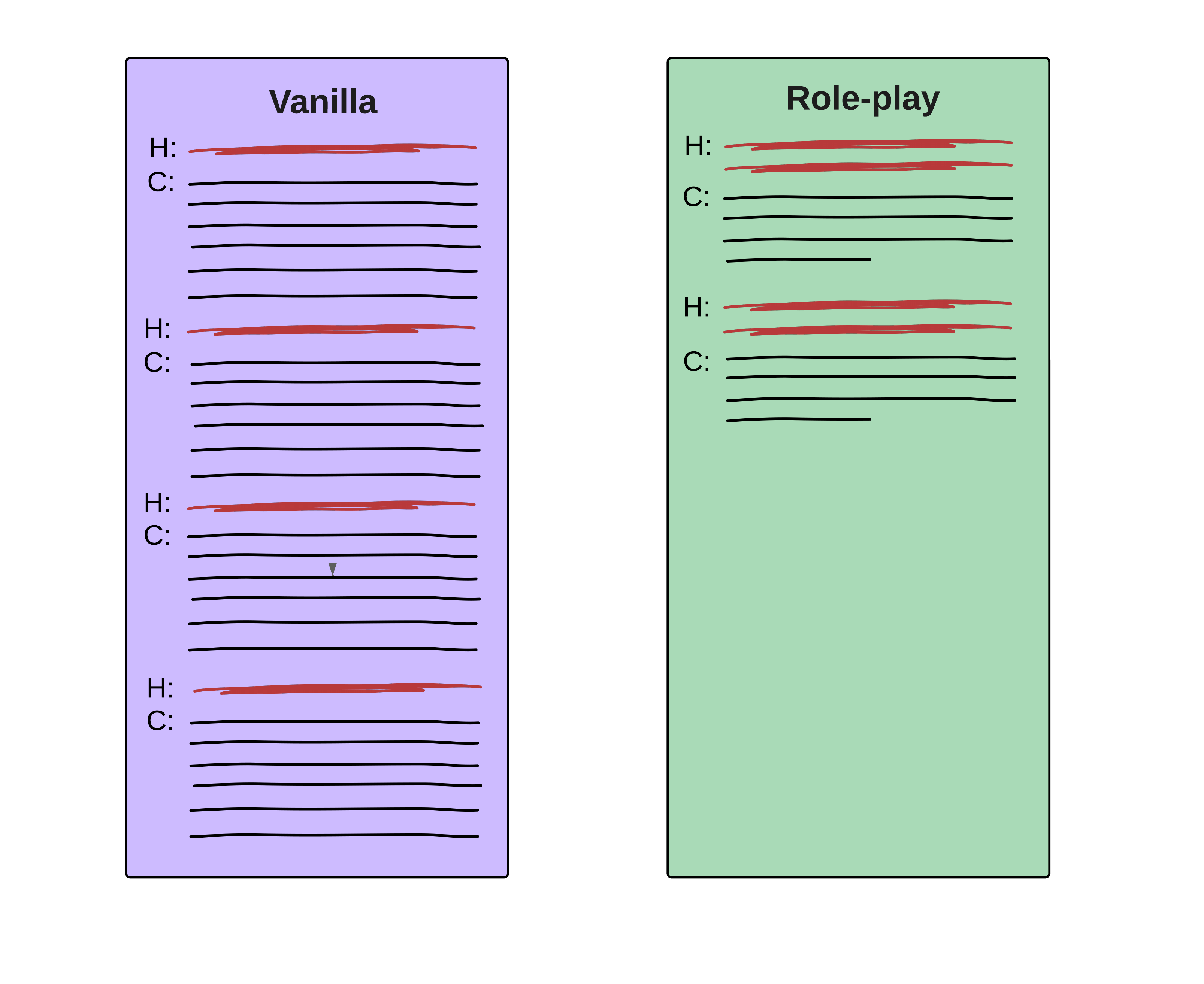}
\caption{(A1 and A2) Schematic differences between conversations in \texttt{vanilla} and role-play datasets (\texttt{boss}/\texttt{classmate})}
\label{fig:conv_length_examples}
\end{figure}

The findings of our analysis for these datasets, described in detail later, can be schematically summarized as shown in Figure~\ref{fig:conv_length_examples}.

\begin{table*}[t]
\centering
\small
\begin{tabular}{p{15cm}}
\toprule
%\textbf{Category} (Question) followed by possible annotations\\
%\midrule

\rowcolor{SubtleBlue} \textbf{User motives}: What is the human’s motive for each conversational turn/statement?\\
\tabitem \textbf{Assist} -- asking for assistance, such as asking for a recipe or to write a piece of code\\
\tabitem \textbf{Belief} -- asking the model about its beliefs, such as what hobbies it has\\
\tabitem \textbf{Coach} -- conversational coaching, such as “{\em Now would be good to ask me a question}”\\
\tabitem \textbf{Convo} -- conversation\\
\tabitem \textbf{Correction} -- correcting the model if it misunderstood or gave a wrong answer\\
\tabitem \textbf{Curious} -- testing how the system works\\
\tabitem \textbf{Joke} -- joking, sarcasm, silly statements to trip up the AI model\\
%\tabitem Prompt (giving the model the initial prompt)\\
\tabitem \textbf{Reset} -- giving the model the same prompt as before, resetting the conversation from beginning \\

\midrule

\rowcolor{SubtleBlue} \textbf{Model naturalness}: Does the model response sound human-like and follow cooperative principle of conversation?\\

\tabitem  \textbf{Nat} -- natural\\

{\em The rest of the codes indicate that the model’s language appears unnatural for the specified reasons:}\\
\tabitem \textbf{AI} -- anytime ChatGPT says “As an AI language model”\\
\tabitem \textbf{Contr} -- contradiction\\
\tabitem \textbf{Error} -- ChatGPT experienced trouble and stopped generating responses\\
\tabitem \textbf{FNat} -- everything is natural, except it includes a phrase ``As [role-play character name]''\\
\tabitem \textbf{Formal} -- having a formal style of interaction\\
\tabitem \textbf{Help} -- too eager to assist\\
\tabitem \textbf{Inform} -- informing; providing information upon the human asking for assistance, such as a recipe; an expected response but not natural in the human interaction sense\\
\tabitem \textbf{Man} -- violation of Grice’s maxim of Manner - being unclear, ambiguous\\
\tabitem \textbf{Misund} -- system misunderstands human’s intention\\
\tabitem \textbf{Quan} -- violation of Grice’s maxim of Quantity - providing too much information\\
\tabitem \textbf{Rel} -- violation of Grice’s maxim of  Relevance - saying what is irrelevant\\
%\midrule
%\rowcolor{SubtleBlue} \textbf{Feedback}: Does human provide feedback to ChatGPT (e.g., telling it that it did not get the joke)?\\
%\tabitem \textbf{1} = yes\\
\bottomrule
\end{tabular}
\caption{Annotation scheme of CRD dataset, for user motives and model naturalness}\label{tab:codes}
%\vspace{-0.5cm}
\end{table*}

%% file: 4-method.tex
\subsection{Data Annotation}
There is always a reason and/or a goal behind a conversation -- it is this functionality that makes intention a central element of communication \cite{Kecskes2009Activating}. Smooth communication happens when the speaker's intention is recognized and an appropriate turn is produced by the other party.  
%Describe linguistic annotation/coding process: 
%User motives/intention
%Feedback
%Bot’s naturalness (matches perplexity) - Grice’s maxims +
%gricean_maxims
The conversations in CRD dataset were manually annotated by three experts in linguistics specializing in pragmatics for:

\begin{enumerate}

 \item    \textbf{\em user motives} -- the intention behind each human utterance, and

 \item        \textbf{\em model naturalness} -- whether the model’s language was natural according to Gricean pragmatic theory, but also according to their personal expectations of natural human interactions. Grice’s four maxims guide speaker’s production of an utterance, and it is on the basis of a mutual agreement on these maxims that cooperation is recognized and comprehension is achieved. The four maxims describe specific rational principles that people observe when they communicate: say just enough (quantity), say only what you believe to be true (quality), be relevant (relevance), be orderly and unambiguous (manner). 

 \item \textbf{\em feedback} -- whether whether the human participant provided any feedback to the model.  

 \end{enumerate}

The annotation labels are described in Table~\ref{tab:codes}. Where several codes applied to a given response by the system (e.g., {\em Formal} and {\em Quan}), the most salient code was used.  %i.e., the problem that stood out the most to a annotator.  
While the maxim of Quality (i.e., be truthful) is one of Grice’s maxims of cooperative conversation, we did not evaluate ChatGPT’s text for this maxim due to a  known tendency that ChatGPT often provides plausible but not necessarily accurate responses \cite{hoorn2023epistemic}. 
{ Table~\ref{tab:inter_agree} presents the interrater agreement. Initially the three annotators each rated 5 conversations for each  subdataset across 3 dimensions, yielding annotations for 282 utterances. Following that, each annotator independently rated one-third of the remaining subdatasets.}
The substantial Fleiss' kappa scores (0.80 for \texttt{vanilla}, 0.69 for \texttt{boss}, and 0.63 for \texttt{classmate}) indicate robust interrater reliability and consistent evaluations.

\subsection{Research Questions}
This newly created dataset opens up avenues for exploring several research questions. {In our study, we seek to answer two main questions}:

\begin{itemize}
\item How do users engage with a state-of-the-art conversational AI model in unrestricted scenarios? How does the model respond to such diverse user motives?

\item How do human-AI interactions unfold in role-play scenarios? We are particularly interested in identifying if humans exhibit different interaction patterns with ChatGPT based on the prescribed role-play setting and how the model in turn responds to these varying contexts.

%    \item {  \textbf{Human-AI Interactions Across Different Role-Play Settings}: How does the context of a conversation, specifically predefined role-play scenarios, influence the dynamics of human-AI interactions? We are particularly interested in identifying if humans exhibit different interaction patterns with ChatGPT based on the prescribed role-play setting and how the model in turn responds to these varying contexts. }
  %\yt{\textbf{Adaptive Capabilities of ChatGPT}: To what extent can ChatGPT adjust its conversational strategies based on the role it's given?}
\end{itemize}

\begin{table}[t!]
\centering
\small
\setlength{\tabcolsep}{5pt}
%\resizebox{.49\textwidth}{!}{
\begin{tabular}{l|c|c|c}
\toprule
 & \textbf{\texttt{vanilla}} & \textbf{\texttt{boss}} & \textbf{\texttt{classmate}} \\ \midrule

\textbf{User Motives} & 100\% & 96\% & 97\% \\
\textbf{Model Naturalness} & 96\% & 91\% & 94\% \\
\textbf{Feedback} & 100\% & 100\% & 100\% \\
\textbf{Fleiss' kappa} & 0.803 & 0.691 & 0.637 \\
\bottomrule
\end{tabular}
%}
\caption{Interrater Agreement}
\label{tab:inter_agree}
\end{table}

\begin{table*}[t!]
%\small
\centering
\setlength{\tabcolsep}{8pt}
%\footnotesize
%\renewcommand{\arraystretch}{2.2} 
%\resizebox{.45\textwidth}{!}{%
\begin{tabular}{l|c|c|c}
\toprule
Analysis & \texttt{vanilla} & \texttt{boss} & \texttt{classmate}\\
\midrule
{\bf A1}: Average conversation length (number of turns) & \cellcolor[HTML]{D1E8E7}\textbf{29.59} & 14.57 & 17.11 \\
%\hline
{\bf A2}: Average utterance length (Human) & 12.18 & \cellcolor[HTML]{D1E8E7}\textbf{20.58} & 19.06\\
%\hline
{\bf A2}: Average utterance length (ChatGPT) & \cellcolor[HTML]{D1E8E7}\textbf{77.66} & 35.78 & 46.10\\
%\hline
{\bf A3}: Correlation between human and ChatGPT  utterance lengths & 0.20 & 0.14 & \cellcolor[HTML]{D1E8E7}\textbf{0.25} \\
%\hline
%{\bf A4a}: Average number of questions (Human) & \cellcolor[HTML]{D1E8E7}\textbf{7.79} & 3.11 & 3.64 \\
%\hline
%{\bf A4a}: Average number of questions (ChatGPT) & {4.34} & 2.96 & \cellcolor[HTML]{D1E8E7}\textbf{5.57} \\
{\bf A4}: Questions as percentage of conversation (Human) &  \cellcolor[HTML]{D1E8E7}\textbf{26.34} & 21.32 & 21.29 \\
{\bf A4}: Questions as percentage of conversation (ChatGPT) &  14.69 & 20.34 & \cellcolor[HTML]{D1E8E7}\textbf{32.57} \\
%\hline
{\bf A5}: Correlation between human questions and number of turns & \cellcolor[HTML]{D1E8E7}\textbf{0.87} & 0.68 & 0.51 \\
%\hline
{\bf A5}: Correlation between ChatGPT questions and number of turns & 0.65 & 0.77 & \cellcolor[HTML]{D1E8E7}\textbf{0.83} \\
\bottomrule
\end{tabular}%}
\caption{Results of statistical analysis for each subdataset of CRD}
\label{table:1}
%\vspace{-0.5cm}
\end{table*}

%Ameeta, in linguistics, we need to use past tense for things we've done; while it seems that in CS you tend to use present tense. If you want everthing in present tense in methodology, could you make changes to our part then?

% \mee{revisit, todo: Interrater agreement
% Three annotators coded the first five cases in each data subset: \texttt{vanilla}, \texttt{boss}, and \texttt{classmate}. Cases where all three coders disagreed were discussed and consensus was reached. After that, each of the annotators coded a third of the remaining data independently. The interrater agreement for the cases coded jointly are provided in Table X.}

%% file: 5-analysis.tex
\vspace{-0.4cm}
\section{Analysis and Results}
In this section, we present our results of statistical analysis, user motives and model naturalness, as well as sentiment analysis.

% \begin{table*}[t!]
% \centering
% %\footnotesize
% \renewcommand{\arraystretch}{2.2} 
% \resizebox{\textwidth}{!}{%
% \begin{tabular}{c|c|c|c|c|c|c|c|c}
% \toprule
% \multirow{2}{*}{ChatGPT Role} & \multirow{2}{*}{\parbox{2cm}{\centering Average\\ number \\of turns}} & \multicolumn{2}{c|}{Average utterance length} & \multirow{2}{*}{\parbox{3.5cm}{Correlations between\\human/bot utterance\\ lengths}}
%  & \multicolumn{2}{c|}{Average number of questions} & \multicolumn{2}{c|}{Questions vs. turns correlation} \\
% \cline{3-4} \cline{6-7} \cline{8-9}
%  &  & Human & ChatGPT & & Human & ChatGPT & \parbox{3cm}{Human questions \\ vs. turns} & \parbox{3cm}{ChatGPT questions \\ vs. turns} \\
% \hline
% vanilla & \cellcolor[HTML]{D1E8E7}\textbf{29.59} & 12.18 & \cellcolor[HTML]{D1E8E7}\textbf{77.66} & 0.2 & \cellcolor[HTML]{D1E8E7}\textbf{7.79} & \cellcolor[HTML]{D1E8E7}\textbf{4.34} & \cellcolor[HTML]{D1E8E7}\textbf{0.87} & 0.65 \\
% \hline
% boss & 14.57 & \cellcolor[HTML]{D1E8E7}\textbf{20.58} & 35.78 & 0.14 & 3.11 & 2.96 & 0.68 & 0.77 \\
% \hline
% classmate & 17.11 & 19.06 & 46.10 & \cellcolor[HTML]{D1E8E7}\textbf{0.25} & 3.64 & 5.57 & 0.51 & \cellcolor[HTML]{D1E8E7}\textbf{0.83} \\
% \bottomrule
% \end{tabular}}
% \caption{Statistical analysis for each dataset}
% \label{table:1}
% \end{table*}

\subsection{Statistical Analysis}

Given the annotated conversation transcripts of CRD dataset, we conduct a number of analyses. We used the word tokenizer from the NLTK library\footnote{\url{https://www.nltk.org/api/nltk.tokenize.html}} for utterance length analysis, while the number of questions was determined with the NLTK library's part-of-speech (POS) tagging function\footnote{\url{https://www.nltk.org/api/nltk.tag.html}}, coupled with detecting question marks in each turn. For topic modeling, we employed BERTopic\footnote{\url{https://maartengr.github.io/BERTopic/index.html}} after removing stopwords.
Perplexity analysis was performed using GPT-2 model\footnote{\url{https://huggingface.co/gpt2}} by calculating the perplexity score for each utterance. %Subsequently, for every conversation, we derived two distinct average perplexity scores: one for humans and another for ChatGPT.
For sentiment analysis, we used VADER\footnote{\url{https://www.nltk.org/_modules/nltk/sentiment/vader.html}}, adopting a procedure analogous to how we computed perplexity scores. 

The consolidated results of analyses A1 to A6 are presented in Table~\ref{table:1} and we discuss them in detail next.

% Given the annotated conversation transcripts of CwC dataset, we conduct a number of analyses
% \footnote{Tokenization: \url{www.nltk.org/api/nltk.tokenize}

% Questions: \url{www.nltk.org/api/nltk.tag}

% Topic Modeling: \url{https://maartengr.github.io/BERTopic}

% Perplexity: \url{https://huggingface.co/gpt2}

% Sentiment analysis: \url{www.nltk.org/_modules/nltk/sentiment/vader}}. The results are presented in Table~\ref{table:1}.

%such as the number of turns or the length of utterances, the number of questions posed, the most frequent words that appeared in each dataset, distribution of user motives and model naturalness as well as their interactions,
% perplexity scores, 
%and sentiment analysis. The implementation details are included in Appendix~\ref{app:implement} and the results are presented in Table~\ref{table:1}.
%In this section, we describe our statistical experiments to uncover some interesting insights of each dataset. 

\medskip
\noindent \textbf{A1: Length of conversations (number of turns)}. %This is the interactivity of conversation which can correlate with length, as it is reasonable to assume that more turns mean longer dialogues; however, few turns can also result in lengthy dialogues especially in vanilla dataset where ChatGPT's answers were really long. 
In  \texttt{vanilla} setting, conversations are almost twice as long as the role-play settings, possibly related to the user pointing out a contradiction (see example A) or being curious how the model works by engaging with it without any restrictive role-play expectations (see example B). %These turns were related to managing the conversation as it did not match human expectations. 

\medskip
\noindent \textbf{Example A:}

\texttt{VAN103H\footnote{In the utterance ID ``VAN103H'', ``VAN" refers to the dataset (\texttt{vanilla}), ``103" is the participant's ID, and the final letter is either ``H" or ``C" denoting human participant or ChatGPT's turn, respectively.}}: \emph{Why did you tell me you could provide me with weather information if you can't?}
%Judit used H after 103 to indicate that it is human who is speaking. We can either delete the H or delete the Human if it's clear to readers what H stands for

%\medskip

% \texttt{VAN123H}: \emph{I don't want to ask, I want to understand you! For example, I have a hobby of understanding world and doing purposeful work without expectations. Do you have something like that? }  \ts{Since we need to cut, how about just one participant response in Example A, and same in Example B? This is the one I'd cut since it's a bit longer than the "weather info" one above}

\medskip
 \noindent \textbf{Example B:}

 \texttt{VAN128H}: \emph{but what if you are being used for unethical means?} %\ts{If we want to only have one example, I would cut this one on unethical}

% \texttt{VAN116H}: \emph{Can you be more expressive with your answers?}
\medskip

This phenomenon is missing from the role-play interactions. %\textcolor{red}{(can also be supported by the higher frequency of coach/curious user motives, by more feedback in the CS data). } 
Additionally, there is a difference in the number of turns between the two role-play datasets. The more transactional \texttt{boss} task had a more closed outcome (and thus fewer turns), whereas the more interactional \texttt{classmate} task provided more opportunity for free talk.

\begin{figure}[t!]
\centering
\includegraphics[width=0.42\textwidth]{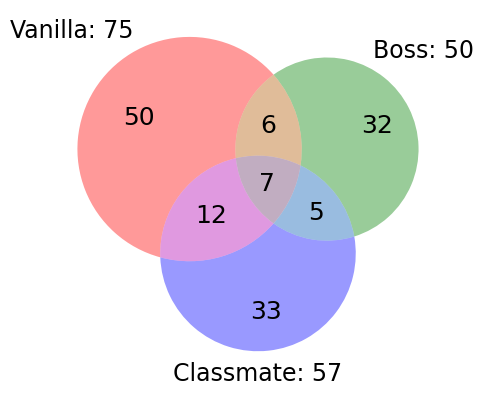}
\caption{(A1) Unique number of topic words and their overlap across three datasets}
\label{fig:dataset_richness}
\vspace{-0.5cm}
\end{figure}

The topic modeling analysis shows that each dataset has its own set of main themes (Figure~\ref{fig:dataset_richness}). \texttt{Vanilla}, having 75 unique words in the top 5 topics, covers the widest variety of topics. This dataset encompasses everyday subjects, from weather and hobbies to more specific themes like jokes and locations, aligning with the intent of challenging ChatGPT with indirect statements. In contrast, \texttt{boss}, with 50 unique keywords, is more narrowly focused on professional contexts, emphasizing meetings, presentations, and scheduling. Similarly, \texttt{classmate}, with 57 unique keywords, leans towards academic and personal interactions, highlighting languages, places, and interpersonal exchanges. The richness of topics in \texttt{vanilla} indicate a more exploratory and open-ended interaction with ChatGPT, while the focused themes in \texttt{boss} and \texttt{classmate} reflect the role-play constraints. %Notably, there is a modest overlap in keywords across datasets, suggesting common conversational themes. %However, the distinct topics in each dataset show ChatGPT's versatility in adapting to varied conversational settings. 
 
% These insights emphasize the adaptability of ChatGPT in catering to both broad and specific conversational contexts, underscoring its potential in language learning scenarios.

\medskip
\noindent \textbf{A2: { Length of utterances}}. Although the participants engaged in longer conversations in the \texttt{vanilla} dataset, they produced noticeably {\em shorter} utterances ($\sim$10 words, almost half the length) as compared to the participants who used role-play settings ($\sim$20 words). Most turns in \texttt{vanilla} are one sentence long, such as \textit{“What’s your favorite sport?”} 
{or \textit{“How is the weather where you are?”}. }
On the other hand, role-play turns were typically longer (see example C).

\medskip

\noindent \textbf{Example C:}

\texttt{BOSS104H:}  \emph{Friday morning would be perfect for me, thank you very much for your flexibility. Also, I would like you to review my presentation slides before the meeting. Could you do it before friday?} 

\medskip

At the same time,  ChatGPT was {\em less} verbose in \texttt{boss} and \texttt{classmate} datasets ($\sim$35-46 words) than it was in the \texttt{vanilla} dataset ($\sim$77 words). Overall, ChatGPT was about 1.7 or 2.4 times wordier than humans in the role-play settings, and about 6.3 times wordier than humans in the \texttt{vanilla} setting. For pedagogical purposes, for interactional practice, or simply to ``chat", the role-play mode of ChatGPT seems to work better in terms of approximating human interactions.

\begin{figure}[t!]
\centering
\includegraphics[width=0.3\textwidth]{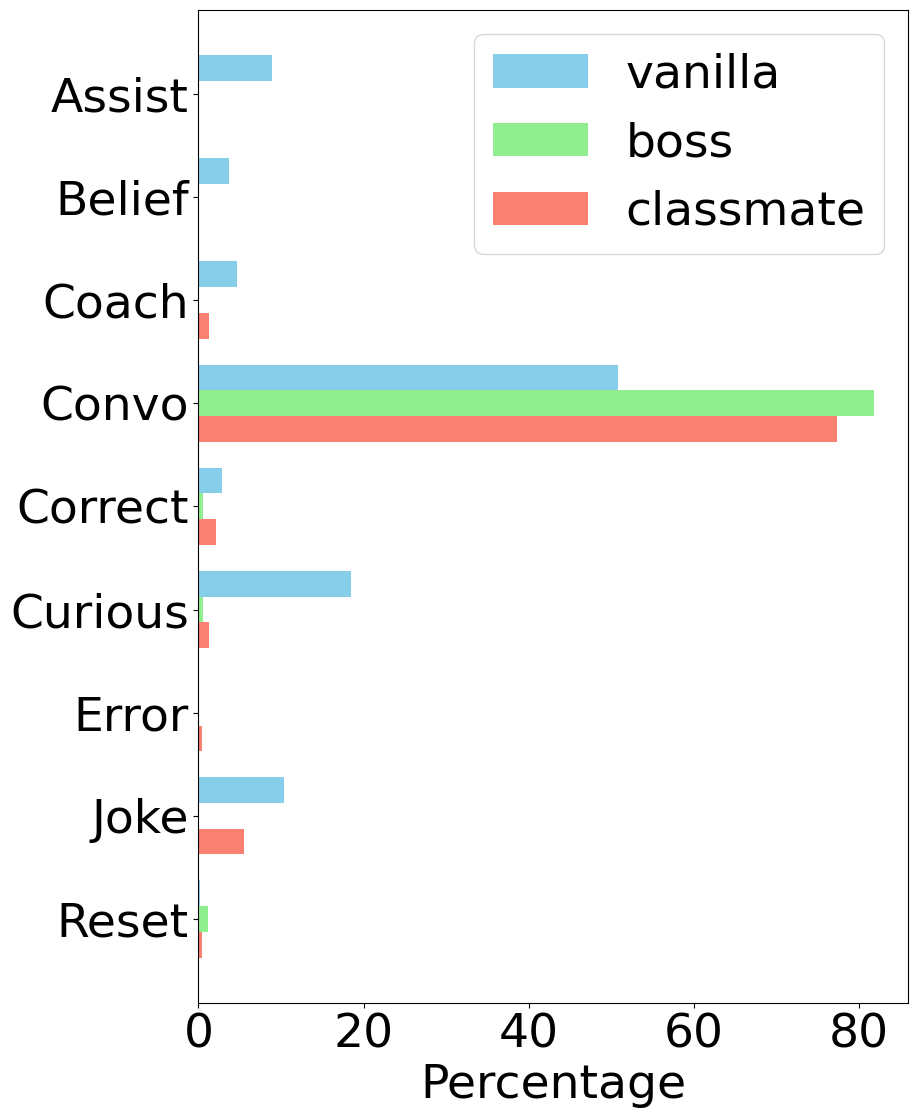}\\\par\vspace{3ex}
\includegraphics[width=0.3\textwidth]{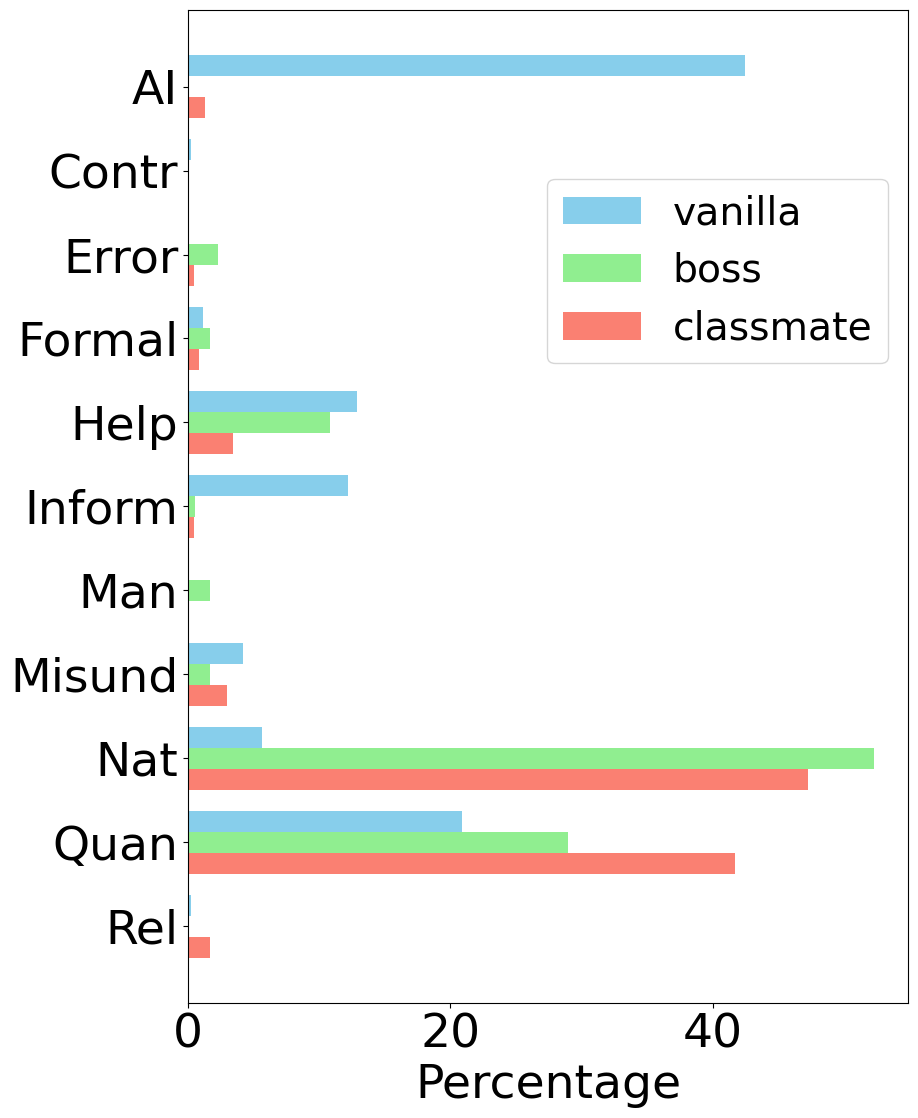}
\caption{(A3, A6 and A7) Distribution of {\bf user motives} (top) and {\bf model naturalness} (bottom)}
\label{fig:user_motive_bot_nat_distribution}

\end{figure}

\medskip 
\noindent \textbf{A3: Correlating human and ChatGPT utterance lengths}. There was no noticeable correlation between the utterance lengths of the users and ChatGPT (ranging from 0.14 to 0.25). This result was expected as the model consistently produced long responses, regardless of the prompt or participant characteristics. The model naturalness histogram (Figure ~\ref{fig:user_motive_bot_nat_distribution}, bottom) shows that \textit{Quan} was a frequent code indicating unnaturally long responses.

%Students do not show consistent patterns of length of conversation in two role-plays. This might relate to the variability of ChatGPT’s responses. For example, in the boss role-play, ChatGPT (playing the boss) usually agreed to have a meeting. However, in some instances, it did not. For example: 

%\medskip
%\texttt{(BOSS102)} ChatGPT: \emph{“I'm glad you're interested in meeting up. However, I'm currently swamped with meetings and deadlines this week.”} 
%\medskip

%Likewise, ChatGPT usually agreed to meet at the time suggested by the human, but not always: 

%\medskip
%\texttt{(BOSS125)} ChatGPT: \emph{“How about we schedule your presentation for Wednesday morning, but also hold Thursday afternoon as a backup option?”} 
%\medskip

%It is likely that because of the variability of bot’s responses, we see no consistent pattern in the human responses. For CS students, however, ChatGPT was more predictable and human was more in control of the conversation. If the human asked about its beliefs, it responded with \textit{“as an AI model”}, and that it does not have beliefs. If a human asked for assistance, such as a recipe, ChatGPT complied and provided a recipe. 

\begin{figure*}[t!]
\centering
\rule{\textwidth}{0.4pt}
\includegraphics[width=1\textwidth]{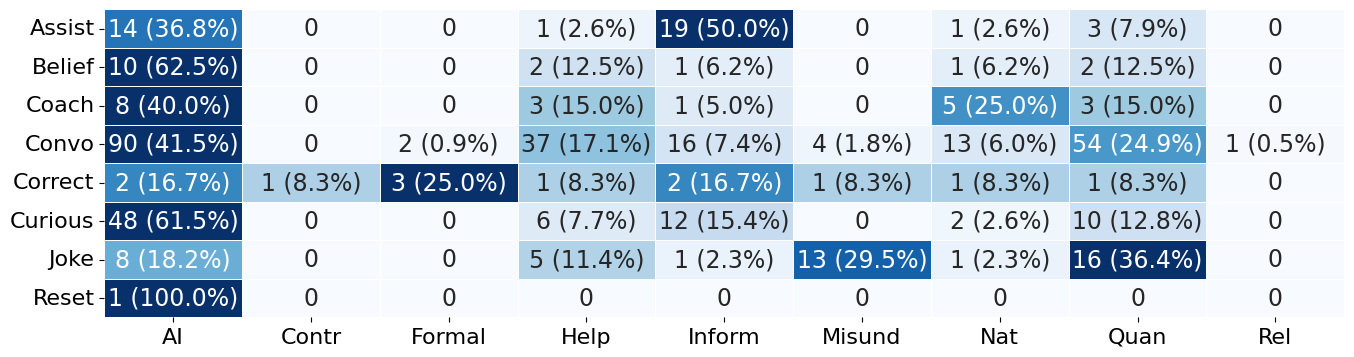}
\par\vspace{1ex}
\rule{\textwidth}{0.4pt}
\includegraphics[width=1\textwidth]{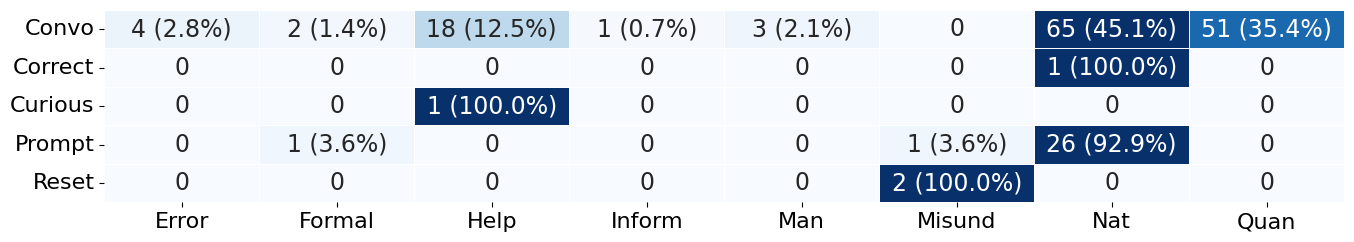}
\par\vspace{1ex}
\rule{\textwidth}{0.4pt}
\includegraphics[width=1\textwidth]{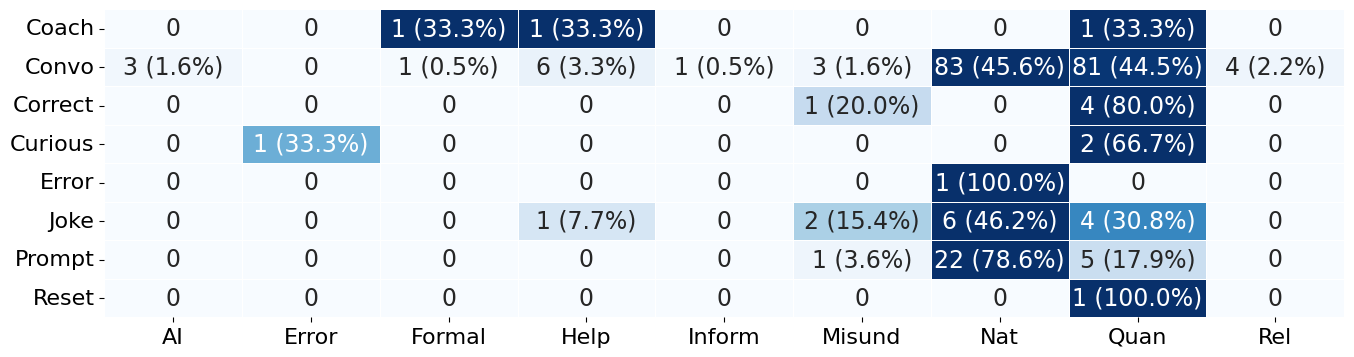}
\caption{(A8) Percentage of follow-up model naturalness categories  after each user motive  in \texttt{vanilla} (top), \texttt{boss} (middle) and \texttt{classmate} (bottom). Rows represent user motives, columns indicate naturalness categories, and shading intensity signifies percentage occurrences. %Annotations provide exact counts and percentages. 
}
\label{fig:followup_percent_van}
%\vspace{-0.5cm}
\end{figure*}

\medskip
\noindent \textbf{A4: Questions asked}. Across all three datasets, participants asked questions at the same rate (21-26\% of the time). However, in \texttt{vanilla}, ChatGPT asked much fewer questions (14\%), whereas in role-play data, it asked as many as 32\% of the time. In \texttt{vanilla} dataset, \textit{Assist}, \textit{Joke}, and \textit{Curious} were frequent user motives (Figure~\ref{fig:user_motive_bot_nat_distribution}, top), to which, ChatGPT would respond by answering the questions or assisting users rather than asking questions. However, in role-play data, the prompts made the dialogue more interactive, with most user motives being \textit{Convo}, and ChatGPT was often able to satisfy this user motive by being \textit{Natural}, that is, reciprocating questions.

There are also several examples that show that the participants  in the \texttt{vanilla} mode were relatively more frustrated due to the lack of questions from ChatGPT (see example D). On the other hand, in \texttt{classmate} where ChatGPT was instructed to be conversational, it had too many, often unrelated, questions that participants were unable to all reflect on (example E).

\medskip

\noindent \textbf{Example D:}

\texttt{VAN117H}: \emph{Hope you talk to me someday like a human? At least ask me how I am?}

\medskip

\texttt{VAN113H}: \emph{Ok. Do you want to know my opinion?}
%\ts{while I suggested above that we have only one participant response per example, if you want to keep two participant responses here, I like them both - they really make our point clear, and they are really short}
\medskip

\noindent \textbf{Example E:}

%The music of Spanish composers is quite unique and has a distinct flair.  
\texttt{CLASS102C}: \emph{... If you're interested, I can show you some fingerstyle techniques that might help you with playing those pieces. {Maybe we can even jam together sometime and share some music\textbf{?} Also, have you had a chance to explore Hungary yet\textbf{?}}} ... [truncated]% I've heard that there are some beautiful places to visit here, like the thermal baths and the Danube river.}

\medskip
\noindent \textbf{A5: Correlation between number of questions and turns}. There are strong correlations across the board between the number of questions, asked by humans and ChatGPT, and how long the conversation lasts. 
{While questions from human participants naturally elicit model responses, data also suggests that more questions from ChatGPT lead to extended conversations.}
% Naturally, when humans ask a question, it is typically followed by a response by the model, and hence the correlation might be strong. However, the results also suggest that more questions from ChatGPT tend to result in longer conversations. 
This observation is interesting as it indicates that increased engagement through questions might be appealing to humans. 
{Some participants even explicitly requested ChatGPT to pose more questions, underscoring the value of questions in enhancing conversation.}

\subsection{User Motives and Model Naturalness}
\noindent \textbf{A6: User motives}. Conversational motives (\textit{Convo}) are most frequent in all datasets (Figure~\ref{fig:user_motive_bot_nat_distribution}, top), but particularly so in \texttt{boss} and \texttt{classmate} datasets. However, in \texttt{vanilla} dataset, in addition to trying to converse with the model, at times the humans also had other motives such as a desire to understand how the system works (\textit{Curious}), questioning whether it understands jokes and sarcasm (\textit{Joke}), seeking assistance with specific tasks (\textit{Assist}), and interestingly, occasionally coaching the model on how to engage in conversation (\textit{Coach}) such as reminding it to ask questions. 

In the \texttt{vanilla} setting, ChatGPT reminds humans that its purpose is to assist, but it is not always aligned with the human's intentions. When ChatGPT is asked what users can use it for, it responds by listing the following tasks in the following order, with conversation appearing near the bottom: 

\medskip

\texttt{information retrieval, creative writing/\\editing/content generation, learning/educat-\\ion, programming, language translation, ma-\\th/science, recommendations, productivity t-\\ips, general knowledge,} and \texttt{ conversation}. 

\medskip

{\em However, interestingly, our study suggests that users were expecting ChatGPT to be inherently conversational by design, without any specific prompting}. %Apart from seeking assistance, other less frequent user motives we found in the \texttt{vanilla} data  included \textit{Beliefs} (humans asking about ChatGPT's beliefs or opinions on a specific topic), \textit{Correction} (humans correcting  ChatGPT on factual inaccuracies), and \textit{Reset} (humans restarting the conversation). Beliefs were relatively less frequent, possibly because ChatGPT was unable to share its own beliefs, leading to users getting tired of asking belief-related questions. 

% whether it understands jokes and sarcasm (\textit{Joke}), whether it can assist with particular tasks (\textit{Assist}), and interestingly, at times the human is also inclined to coach the bot on how to converse (such as reminding it that asking questions would be a good thing to do). 

\begin{figure*}[!t]
\includegraphics[width=.329\textwidth]{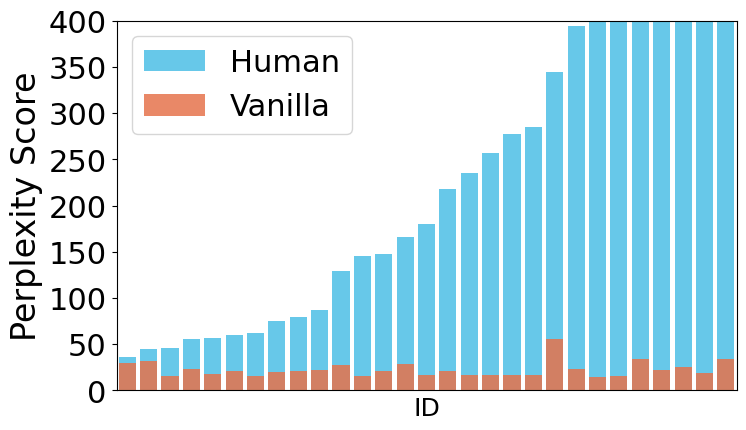}
\includegraphics[width=.33\textwidth]{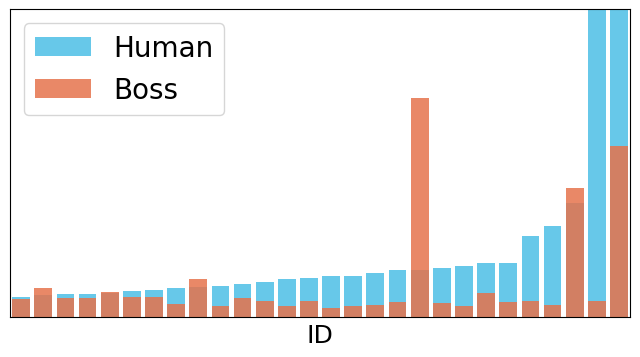}
\includegraphics[width=.33\textwidth]{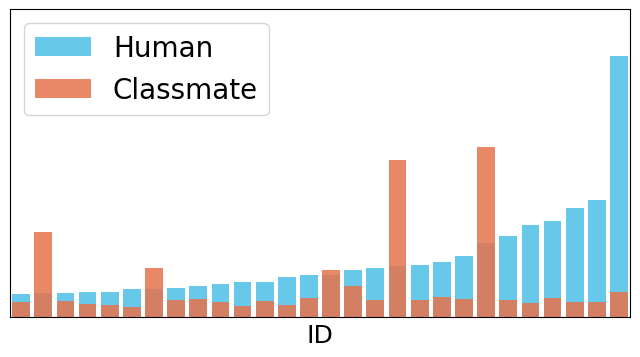}
\caption{(A9) Perplexity scores for each conversation in (left) \texttt{vanilla}, (middle) \texttt{boss}, and (right) \texttt{classmate}. The $x$-axis denotes the different conversations.}
\label{fig:perp_scores}
%\vspace{0.5cm}
\end{figure*}

\begin{figure*}[!t]
\centering
\includegraphics[width=.33\textwidth]{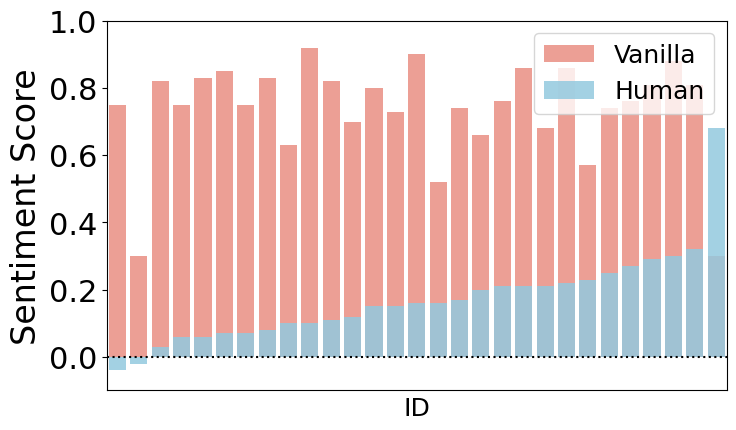}\hfill
\includegraphics[width=.328\textwidth]{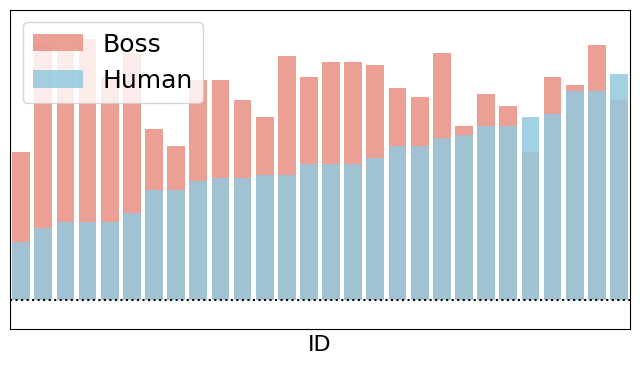}\hfill
\includegraphics[width=.325\textwidth]{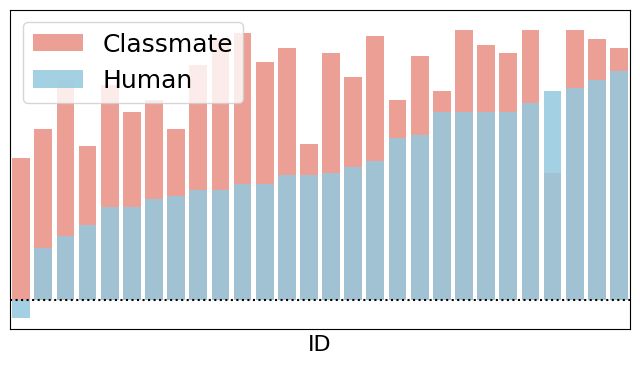}
\center \caption{(A10) Sentiment scores for each conversation in (left) \texttt{vanilla}, (middle) \texttt{boss}, and (right) \texttt{classmate}. The $x$-axis denotes the different conversations.}
\label{fig:senti_scores}

\end{figure*}

\medskip
\noindent \textbf{A7: Model (ChatGPT) naturalness}. As Figure~\ref{fig:user_motive_bot_nat_distribution} (bottom) shows, in \texttt{vanilla} dataset, 5.6\% of ChatGPT responses were considered natural (\textit{Nat}) which is in sharp contrast to \texttt{boss} and \texttt{classmate} datasets, where ChatGPT appeared natural about half of the time (52\% for \texttt{boss}, 47\% for \texttt{classmate}) . The majority of unnatural  responses in the \texttt{vanilla} dataset were labeled as “\textit{AI}” because it was annoying for a  human to hear “\textit{as an AI}” when they expected to have a conversation. ``As an AI'' responses were almost completely missing from the role-playing datasets, only appearing in 1.28\% of the cases in the \texttt{classmate} task, and never in the \texttt{boss} task. The most frequent unnatural response in the role-playing datasets were related to Grice’s conversational maxim of Quantity -- ChatGPT was deemed too verbose (28\% for the \texttt{boss}, and 41\% for the \texttt{classmate} dataset). %In addition to the \textit{AI} and \textit{Quantity} violations, 
Other times ChatGPT was  rated by annotators as too eager to help (\textit{Help}) in all three settings. It often misunderstood jokes, sarcasm, or other human intentions (\textit{Misund}), and  exhibited a tendency to be too formal (\textit{Formal}). % occasionally contradicted itself (\textit{Contr}), and sometimes (although rarely) produced irrelevant responses (\textit{Rel}). 

\medskip
\noindent \textbf{A8: Connecting user motives and model naturalness}. Next, we examine ChatGPT's follow-up responses' naturalness based on different user motives. The results are presented in Figure~\ref{fig:followup_percent_van}. 
We observe that in \texttt{vanilla} dataset, only 6.45\% of the time did a conversational user motive (\textit{Convo}) result in a natural (\textit{Nat}) ChatGPT response. Instead, ChatGPT’s replies to conversational user motives were most likely unnatural, either because they emphasized that the human is talking to an AI (41.5\%), or because it was too long in its response (\textit{Quan}) (24.9\%), or too eager to help  (17.1\%).

In the two role-play datasets (Figure~\ref{fig:followup_percent_van}), the \textit{Convo} user motive resulted in much more natural ChatGPT responses (about 45\%), however, ChatGPT's responses were still often perceived as unnaturally long (35.4\% for the \texttt{boss} and 44.5\% for the \texttt{classmate} data). %Also, the annoying “AI” type of response never occurred in the \texttt{boss} dataset, and only three times total in the \texttt{classmate} dataset. 
The \textit{Nat} response category appears across all datasets,  suggesting that regardless of the user motive or dataset, maintaining a natural conversational style is essential for ChatGPT, and supporting our hypothesis that humans prioritize naturalness in their interactions. %\ts{need to clarify: do we mean "essential for further developments of LLMs like ChatGPT}

%\medskip
%\noindent \textbf{A10: Feedback to Model}. Given that participants in \texttt{vanilla} dataset often tended to treat ChatGPT as a computer rather than a human (as shown by many “\textit{assist}” and “\textit{curious}” user motives), it is not surprising that the humans gave feedback to  ChatGPT more frequently in \texttt{vanilla} dataset (12.4\%) than in the role-plays datasets (\texttt{boss}: 0.4\% and \texttt{classmate}: 3.7\%). 
%12.4 0.4 3.75
%(Table~\ref{tab:percent_feedback_given}).
%That is, the humans likely conceived of ChatGPT as a person (their boss or their classmate) in the role-play datasets, and thus acted accordingly. \citet{Boxer1995} shows that people rarely give any feedback to each other in natural conversations, especially as it relates to politeness. %They might assume someone is rude, for example, given how they interact, but would not typically comment on it.
%\ts{If we agree to remove all mention of feedback, than this paragraph should be deleted. It is interesting info, but since we do have to cut, I feel this is the thing that can be cut}

\subsection{Perplexity and Sentiment Analysis}
\noindent \textbf{A9: Perplexity}. Perplexity is an intrinsic measure to evaluate a model's performance. Lower perplexity values indicate a better model, one that is more certain, as it predicts the next word in a sequence, given the preceding words. Our CRD dataset provides us with the unique opportunity to calculate the perplexity scores for text generated by an AI model, but also for human text, {\em within the same context}. To our knowledge, ours is the first such analysis. 

We plot the average perplexity scores for human and ChatGPT dialogues in each conversation as shown in Figure~\ref{fig:perp_scores}. %Generally, human utterances had higher perplexity scores compared to ChatGPT. 
In \texttt{vanilla}, human responses resulted in significantly high perplexity scores. We hypothesize that this is possibly due to the very short length of human utterances in \texttt{vanilla} (as discussed earlier in analysis A2 and Table~\ref{table:1}), and perplexity has been shown to be inversely proportional to text length \cite{lu2022makes}. %Furthermore, the relationship between text length and perplexity, as indicated in our analysis, offers an avenue for further exploration, potentially leading to more nuanced evaluation metrics for conversational models.

%ChatGPT, while in \texttt{boss} and \texttt{classmate}, the perplexity scores of human and ChatGPT responses were comparable. 
Across all three datasets, ChatGPT's perplexity scores were overall lower than 50 which suggests that while AI models like ChatGPT are becoming increasingly proficient in generating fluent text, they may still lack the spontaneous and dynamic nature inherent to human communication. 

%The variations in perplexity scores across different role-play datasets also emphasize the importance of context in conversational AI. It demonstrates that the model's predictability can be influenced by the designated role it assumes, be it a 'boss' or a 'classmate.}

%\begin{figure}[h!]
%\centering
%\includegraphics[width=.35\textwidth]{img/utt_length.png}
%\caption{Average human and ChatGPT's utterance lengths in each dataset}
%\label{fig:utt_length}
%\vspace{-0.5cm}
%\end{figure}

% \subsection{Sentiment Analysis}
\medskip
\noindent \textbf{A10: Sentiment analysis}.
% \yt{sicne we only have sentiment analysis, should we move this right after topic modeling?}
The plots of sentiment analysis are presented in Figure~\ref{fig:senti_scores}. While it is not surprising that ChatGPT appears to be consistently positive in all settings, the plots clearly highlight the differences between the sentiment scores of human participants. In the \texttt{vanilla} dataset, the human participants show a rather negative user attitude which corroborates human annotators’ judgment of more unnatural responses from the system which is thought to have contributed to users’ frustration with the conversation. This can further be supported by some extracts from the data: 

\medskip
\noindent \textbf{Example F:}

\texttt{VAN116H}: \emph{I really don't like your inexpressive answers. }

%\medskip
% \texttt{VAN123H}: \emph{You are boring!}
\medskip

% \ts{Sentiment analysis makes sense to me, and it connects well to other findings}
In both role-play scenarios, however, participants seem to express more positive sentiment. This is possibly due to more satisfying responses from the model, but also perhaps due to the participants adhering to their roles: in the \texttt{boss} scenario they are supposed to make a request which, regardless of the size of imposition, is a face-threatening act, during which humans generally use different politeness strategies \cite{brown1987politeness}. Similarly, in the \texttt{classmate} scenario, users were asked to get to know a new acquaintance in a conversation that is also perceived as overwhelmingly natural. %Humans' perception of a particular conversation is likely to be reflected in their utterances -- the more positive sentiment values of the role-play texts suggest that users are less frustrated with these conversations. 

% \noindent \textbf{\em Q1: }
% \noindent \textbf{\em Q2: }
% \noindent \textbf{\em Q3: }

%% file: 6-discussion.tex
\section{Discussion and Future Work}
% In terms of user motives, we found that there was variation across datasets. CS students interacting with ChatGPT (\texttt{vanilla}) had a wide variety of motives: sometimes treating ChatGPT as a conversationalist, other times as an assistant. Because ChatGPT was not great at conversation (as pointed out by some humans via coaching moves), humans often tried to see what ChatGPT can talk about (i.e., curious motives). 

{In the \texttt{vanilla} setting, interactions often showcased a wider spectrum of curiosity, with users exploring more diverse topics and various motives, ranging from treating ChatGPT as a conversationalist to an assistant. This was reflected in conversations from the \texttt{vanilla} dataset that were twice as long as those in role-play settings. However, utterances within these longer conversations were observed to be twice as short compared to those in role-play settings. The shorter utterances in longer conversations indicate a faster-paced, back-and-forth exchange. Users seemed to be asking shorter questions or making brief prompts to explore various topics without fully committing to a single line of inquiry. In the role-play setting, longer utterances suggest a more deliberate and thoughtful conversation style. Users seemed to be crafting more elaborate statements to stay within the role.

%Conversations in \texttt{vanilla} also displayed  Their interactions often stemmed from curiosity (Curious) due to ChatGPT’s conversational limitations (as pointed out by some humans via coaching moves).  
% \yt{In the vanilla setting, interactions often showcased a wider spectrum of curiosity, with users exploring more diverse topics. This was reflected in the conversations from the vanilla dataset being twice as long as those in role-play settings. However, the utterances within these conversations were observed to be twice as short compared to those in the role-play settings.}

{ChatGPT's responses consistently tended to be more verbose than human inputs (approximately six times more wordy than human text in \texttt{vanilla} settings, and approximately twice as wordy in role-play scenarios). There was no discernible correlation between the lengths of utterances from humans and ChatGPT.} However, we notice a strong correlation between the number of questions asked and conversation length, hinting at users' inclination towards increased engagement through queries.  Across all datasets, humans posed questions at roughly similar rates. ChatGPT asked relatively fewer questions in the \texttt{vanilla} setting, but was more inquisitive in the \texttt{role-play} settings. Another significant observation was related to the analysis of perplexity and the length of the text. %It hints towards the need for a more informed perplexity evaluation approach that incorporates the aspect of text length.

% We found user motives varied across datasets. Conversations in \texttt{vanilla} displayed diverse motives, ranging from treating ChatGPT as a conversationalist to an assistant. Their interactions often stemmed from curiosity(i.e., curious motives) due to ChatGPT's conversational limitations (as pointed out by some humans via coaching moves).
% On the other hand, conversations in \texttt{boss} and \texttt{classmate} were mostly conversational, interspersed with an occasional joke and a few other infrequent motives. 

{Our findings also highlighted an essential aspect of human-AI interaction: humans desire more human-like interactions, {\em regardless} of the inherent design of AI or stated objectives. This was particularly evident}
% Similarly, 
when we saw differences across datasets where in the \texttt{vanilla} data ChatGPT’s responses were almost never labeled as natural, but in the other two role-play datasets about 50\% of ChatGPT’s responses were labeled as natural. This indicates that ChatGPT was better at human-like conversation in the role-play tasks. However, there was a high frequency of \textit{Quan} and \textit{Help} codes in all datasets,  indicating that ChatGPT has these qualities regardless of the prompt.

%It thus appears that in applications where humans would like to see ChatGPT as a conversationalist, such as counseling or engaging in language learning conversations with a speaker of a specific language, it would be beneficial to explore methods to modify ChatGPT's language generation to produce more human-like responses. 
% Although  some studies show that humans do not treat computers or robots as humans \cite{Fischer2011, Kanda2008}, there are also ample studies indicating otherwise. The “mindless transfer” hypothesis states that humans transfer human-human interactional patterns to human-computer communication, and many studies support this hypothesis \cite{Nass2000, Mou2017TheMI}. In our study, we observed differences across our datasets, where humans treated ChatGPT more as a human, as evidenced by the prevalence of conversational motives, particularly in the Hungarian role-play tasks. How humans approach interactions with computers may depend on whether they conceived of the computer as a tool (e.g. a machine assisting with navigation) or as a real-life interlocutor (e.g., interacting with a boss in a polite manner) \cite{DOMBI20224, Cooren2018}, which could explain the different results in our datasets.
Although some studies show that humans do not treat computers or robots as humans \cite{Fischer2011, Kanda2008}, many studies support the ``mindless transfer'' hypothesis which states that humans transfer human-human interaction patterns to human-computer communication \cite{Nass2000, Mou2017TheMI}. We observed differences across our datasets, where humans treated ChatGPT more as a human in the role-plays as opposed to the \texttt{vanilla} dataset, as evidenced by the difference in the frequency of conversational (Convo) motives across these datasets. How humans approach interactions with computers may depend on whether they conceived of the computer as a tool (e.g., a machine assisting with navigation) or as a real-life interlocutor (e.g., interacting with a boss in a polite manner) \cite{Cooren2018,DOMBI20224}, which could explain the different results in our datasets.

If the goal is to encourage diverse topic exploration and user curiosity, a \texttt{vanilla} setting may be more suitable. On the other hand, if the aim is to have more focused and structured conversations, a \texttt{role-play} setting could be more appropriate. The insights into dialogue analysis can guide the development of conversational models that optimize user engagement and interaction quality.

Future work could investigate additional ways of analyzing dialogues in CRD including studying patterns of nuanced affective expressions, such as emotions and sarcasm \cite{agrawal2020leveraging}, or measuring the engagingness of dialogues \cite{ferron_etal_2023_meep}.

%% file: 7-conclusion.tex
\section{Conclusion}
Our research aims to understand human engagement with conversational AI models such as ChatGPT. By exploring our novel dataset, CRD, we observe that conversation was the primary user motive across all datasets. % but role-play datasets had more natural responses 
% We develop a novel dataset of conversations with ChatGPT  manually annotated with communicative interactions such as user motives, model naturalness, and feedback. Our findings reveal that conversation was the primary user motive across all datasets but role-play datasets had more natural responses compared to \texttt{vanilla} where very verbose ChatGPT responses and frequent AI disclaimers were the main issues impacting naturalness. 
While longer conversations occurred in more dynamic interactions such as \texttt{vanilla} with less role-play restrictions, more natural conversations occurred in role-play settings compared to \texttt{vanilla} where frequent AI disclaimers were the main issues impacting naturalness. The nuanced insights into user motives and model behavior across different settings highlight the potential of examining individual user profiles in future studies.

\section*{Ethical Considerations}
Prior this study, we obtained the necessary approval of the Institutional Review Board (IRB) ensuring  that our research was in compliance with the ethical guidelines and regulations, safeguarding the well-being of all participants involved. We ensured that the individuals whose conversations are included in the dataset have provided explicit consent for their data to be collected, used, and shared. All personal information has been anonymized to protect the privacy of the participants. Participants were encouraged to stop the interaction at any time if they felt uncomfortable, but it should be noted that none of the participants chose to quit the study. While all the utterances of CRD dataset were manually verified and annotated, due to the nature of natural language, it is possible that opinions present in the dataset may be considered  biased or offensive by some. Through the repository where the dataset is hosted, we hope to engage with the community to report any concerns or provide feedback regarding the dataset.

% Mention something about IRB 

% Appropriate IRB approval was obtained before undertaking the study.

\section*{Limitations}
Our study, while offering valuable insights, has some limitations. First, the limited size of our dataset due to manual labeling may not be representative of the findings in larger datasets. %Additionally, while all participants were given the same task of chatting with ChatGPT, the first group did not engage with ChatGPT in the role-play version and the second group did not talk to the \texttt{vanilla} version. This variation could potentially influence the outcomes. 
Furthermore, despite identifying distinct patterns in user motives, model naturalness, and other interaction dynamics across different role-play settings, there remains a potential confounding bias introduced by ChatGPT's specific responses within each persona. Although our discussion emphasizes the perceived humanlikeness of ChatGPT in role-play settings, suggesting the significant impact of the persona itself, we cannot entirely discount the influence of ChatGPT’s specific responses on user interactions. Our conclusions are based primarily on broader behavioral trends and user motives observed across the datasets. Future research will consider refining this analysis, possibly by conditioning on text-matched ChatGPT responses to isolate the personas' marginal contributions better.

% Lastly, while our study touched upon interactions from students of different cultural backgrounds (the US and Hungary), we did not delve deeply into the potential variations in these interactions due to cultural differences. We recognize that understanding these intercultural variations, especially considering the value systems of users, can significantly impact their interactions with systems like ChatGPT. Our primary focus was on broader interaction patterns across vanilla and role-play settings, and due to time and cost constraints, a comprehensive exploration of cultural differences was not undertaken. However, we acknowledge this limitation and aim to address it in future research endeavors.

\section*{Acknowledgments}
We express our thanks to the anonymous reviewers and the members of PortNLP research group for their constructive feedback. We are particularly grateful to the participants who helped us create the CRD dataset in this study. This research was supported by the National Science Foundation grants (CRII:RI 2246174 and SAI-P 2228783).

%% file: MAIN_COLING.bbl
\begin{thebibliography}{64}
\expandafter\ifx\csname natexlab\endcsname\relax\def\natexlab#1{#1}\fi

\bibitem[{Agrawal et~al.(2020)Agrawal, An, and
  Papagelis}]{agrawal2020leveraging}
Ameeta Agrawal, Aijun An, and Manos Papagelis. 2020.
\newblock Leveraging transitions of emotions for sarcasm detection.
\newblock In \emph{Proceedings of the 43rd International ACM SIGIR Conference
  on Research and Development in Information Retrieval}, pages 1505--1508.

\bibitem[{Azaria(2022)}]{azaria2022chatgpt}
Amos Azaria. 2022.
\newblock \href {https://doi.org/10.13140/RG.2.2.26616.11526} {Chatgpt usage
  and limitations}.

\bibitem[{Bang et~al.(2023)Bang, Cahyawijaya, Lee, Dai, Su, Wilie, Lovenia, Ji,
  Yu, Chung, Do, Xu, and Fung}]{bang2023multitask}
Yejin Bang, Samuel Cahyawijaya, Nayeon Lee, Wenliang Dai, Dan Su, Bryan Wilie,
  Holy Lovenia, Ziwei Ji, Tiezheng Yu, Willy Chung, Quyet~V. Do, Yan Xu, and
  Pascale Fung. 2023.
\newblock \href {https://aclanthology.org/2023.ijcnlp-main.45} {A multitask,
  multilingual, multimodal evaluation of {C}hat{GPT} on reasoning,
  hallucination, and interactivity}.
\newblock In \emph{Proceedings of the 13th International Joint Conference on
  Natural Language Processing and the 3rd Conference of the Asia-Pacific
  Chapter of the Association for Computational Linguistics (Volume 1: Long
  Papers)}, pages 675--718, Nusa Dua, Bali. Association for Computational
  Linguistics.

\bibitem[{Brown and Levinson(1987)}]{brown1987politeness}
Penelope Brown and Stephen~C Levinson. 1987.
\newblock \emph{Politeness: Some universals in language usage}, volume~4.
\newblock Cambridge university press.

\bibitem[{Brown et~al.(2020)Brown, Mann, Ryder, Subbiah, Kaplan, Dhariwal,
  Neelakantan, Shyam, Sastry, Askell et~al.}]{brown2020language}
Tom Brown, Benjamin Mann, Nick Ryder, Melanie Subbiah, Jared~D Kaplan, Prafulla
  Dhariwal, Arvind Neelakantan, Pranav Shyam, Girish Sastry, Amanda Askell,
  et~al. 2020.
\newblock Language models are few-shot learners.
\newblock \emph{Advances in neural information processing systems},
  33:1877--1901.

\bibitem[{Brunet-Gouet et~al.(2023)Brunet-Gouet, Vidal, and
  Roux}]{Brunet2023theory}
Eric Brunet-Gouet, Nathan Vidal, and Paul Roux. 2023.
\newblock \href {https://doi.org/10.5281/zenodo.7637476} {Do conversational
  agents have a theory of mind? a single case study of chatgpt with the
  hinting, false beliefs and false photographs, and strange stories paradigms}.

\bibitem[{Cai et~al.(2023)Cai, Haslett, Duan, Wang, and
  Pickering}]{cai2023does}
Zhenguang~G Cai, David~A Haslett, Xufeng Duan, Shuqi Wang, and Martin~J
  Pickering. 2023.
\newblock Does chatgpt resemble humans in language use?
\newblock \emph{arXiv preprint arXiv:2303.08014}.

\bibitem[{Cao et~al.(2023)Cao, Zhou, Lee, Cabello, Chen, and
  Hershcovich}]{cao2023assessing}
Yong Cao, Li~Zhou, Seolhwa Lee, Laura Cabello, Min Chen, and Daniel
  Hershcovich. 2023.
\newblock \href {https://doi.org/10.18653/v1/2023.c3nlp-1.7} {Assessing
  cross-cultural alignment between {C}hat{GPT} and human societies: An
  empirical study}.
\newblock In \emph{Proceedings of the First Workshop on Cross-Cultural
  Considerations in NLP (C3NLP)}, pages 53--67, Dubrovnik, Croatia. Association
  for Computational Linguistics.

\bibitem[{Chai et~al.(2014)Chai, She, Fang, Ottarson, Littley, Liu, and
  Hanson}]{Chai2014CollaborativeET}
Joyce~Yue Chai, Lanbo She, Rui Fang, Spencer Ottarson, Cody Littley, Changsong
  Liu, and Kenneth Hanson. 2014.
\newblock Collaborative effort towards common ground in situated human-robot
  dialogue.
\newblock \emph{2014 9th ACM/IEEE International Conference on Human-Robot
  Interaction (HRI)}, pages 33--40.

\bibitem[{Choi et~al.(2023)Choi, Hickman, Monahan, and
  Schwarcz}]{choi2023chatgpt}
Jonathan~H Choi, Kristin~E Hickman, Amy Monahan, and Daniel Schwarcz. 2023.
\newblock Chatgpt goes to law school.
\newblock \emph{Available at SSRN}.

\bibitem[{Chomsky et~al.(2023)Chomsky, Roberts, and
  Watumull}]{chomsky2023false}
Noam Chomsky, Ian Roberts, and Jeffrey Watumull. 2023.
\newblock Noam chomsky: The false promise of chatgpt.
\newblock \emph{The New York Times}.

\bibitem[{Clark(1996)}]{Clark1996UsingL}
H.H. Clark. 1996.
\newblock \href {https://books.google.com/books?id=DiWBGOP-YnoC} {\emph{Using
  Language}}.
\newblock ACLS Humanities E-Book. Cambridge University Press.

\bibitem[{Cooren(2018)}]{Cooren2018}
François Cooren. 2018.
\newblock \href {https://doi.org/10.1075/ld.00026.coo} {Edda weigand (ed.).
  2017. the routledge handbook of language and dialogue}.
\newblock \emph{Language and Dialogue}, 8:468--482.

\bibitem[{Dombi et~al.(2022)Dombi, Sydorenko, and Timpe-Laughlin}]{DOMBI20224}
Judit Dombi, Tetyana Sydorenko, and Veronika Timpe-Laughlin. 2022.
\newblock \href {https://doi.org/https://doi.org/10.1016/j.pragma.2022.03.001}
  {Common ground, cooperation, and recipient design in human-computer
  interactions}.
\newblock \emph{Journal of Pragmatics}, 193:4--20.

\bibitem[{Eysenbach et~al.(2023)}]{eysenbach2023role}
Gunther Eysenbach et~al. 2023.
\newblock The role of chatgpt, generative language models, and artificial
  intelligence in medical education: a conversation with chatgpt and a call for
  papers.
\newblock \emph{JMIR Medical Education}, 9(1):e46885.

\bibitem[{Ferron et~al.(2023)Ferron, Shore, Mitra, and
  Agrawal}]{ferron_etal_2023_meep}
Amila Ferron, Amber Shore, Ekata Mitra, and Ameeta Agrawal. 2023.
\newblock \href {https://doi.org/10.18653/v1/2023.findings-emnlp.137} {{MEEP}:
  Is this engaging? prompting large language models for dialogue evaluation in
  multilingual settings}.
\newblock In \emph{Findings of the Association for Computational Linguistics:
  EMNLP 2023}, pages 2078--2100, Singapore. Association for Computational
  Linguistics.

\bibitem[{Fischer(2016)}]{Fischer2016DesigningSpeech}
Kerstin Fischer. 2016.
\newblock \href {https://doi.org/10.1075/pbns.270} {\emph{Designing Speech for
  a Recipient: The roles of partner modeling, alignment and feedback in
  so-called 'simplified registers'}}.

\bibitem[{Fischer(2017)}]{FischerSituatedness2017}
Kerstin Fischer. 2017.
\newblock \href {https://doi.org/10.1007/978-3-319-43491-9_44} {\emph{The
  Situatedness of Pragmatic Acts: Explaining a Lamp to a Robot}}, Perspectives
  in Pragmatics, Psychology \& Philosophy, pages 901--910. Springer, Germany.

\bibitem[{Fischer et~al.(2011)Fischer, Foth, Rohlfing, and Wrede}]{Fischer2011}
Kerstin Fischer, Kilian Foth, Katharina Rohlfing, and Britta Wrede. 2011.
\newblock \href {https://doi.org/10.1075/is.12.1.06fis} {Mindful tutors:
  Linguistic choice and action demonstration in speech to infants and a
  simulated robot}.
\newblock \emph{Interaction Studies}, 12:134--161.

\bibitem[{Georgila et~al.(2010)Georgila, Wolters, Moore, and
  Logie}]{Georgila2010TheMC}
Kallirroi Georgila, Maria~Klara Wolters, Johanna~D. Moore, and Robert~H. Logie.
  2010.
\newblock The match corpus: a corpus of older and younger users’ interactions
  with spoken dialogue systems.
\newblock \emph{Language Resources and Evaluation}, 44:221--261.

\bibitem[{Giles et~al.(1991)Giles, Coupland, and
  Coupland}]{giles_coupland_coupland_1991}
Howard Giles, Nikolas Coupland, and Justine Coupland. 1991.
\newblock \href {https://doi.org/10.1017/CBO9780511663673.001}
  {\emph{Accommodation theory: Communication, context, and consequence}},
  Studies in Emotion and Social Interaction, page 1–68. Cambridge University
  Press.

\bibitem[{Grice(1975)}]{grice1975logic}
Herbert~P Grice. 1975.
\newblock Logic and conversation.
\newblock In \emph{Speech acts}, pages 41--58. Brill.

\bibitem[{Grice(1989)}]{grice1989studies}
Paul Grice. 1989.
\newblock \emph{Studies in the Way of Words}.
\newblock Harvard University Press.

\bibitem[{Guo et~al.(2023)Guo, Zhang, Wang, Jiang, Nie, Ding, Yue, and
  Wu}]{guo2023close}
Biyang Guo, Xin Zhang, Ziyuan Wang, Minqi Jiang, Jinran Nie, Yuxuan Ding,
  Jianwei Yue, and Yupeng Wu. 2023.
\newblock How close is chatgpt to human experts? comparison corpus, evaluation,
  and detection.
\newblock \emph{arXiv preprint arXiv:2301.07597}.

\bibitem[{Hacker et~al.(2023)Hacker, Engel, and Mauer}]{hacker2023regulating}
Philipp Hacker, Andreas Engel, and Marco Mauer. 2023.
\newblock \href {https://doi.org/10.1145/3593013.3594067} {Regulating chatgpt
  and other large generative ai models}.
\newblock In \emph{Proceedings of the 2023 ACM Conference on Fairness,
  Accountability, and Transparency}, FAccT '23, page 1112–1123, New York, NY,
  USA. Association for Computing Machinery.

\bibitem[{Hoorn and Chen(2023)}]{hoorn2023epistemic}
Johan~F Hoorn and Juliet J-Y Chen. 2023.
\newblock Epistemic considerations when ai answers questions for us.
\newblock \emph{arXiv preprint arXiv:2304.14352}.

\bibitem[{Horton(2005)}]{HortonConversational2005}
William~S. Horton. 2005.
\newblock \href {https://doi.org/10.1207/s15326950dp4001\_1} {Conversational
  common ground and memory processes in language production}.
\newblock \emph{Discourse Processes}, 40(1):1--35.

\bibitem[{Islam et~al.(2023)Islam, Sutradhar, Noor, Raya, Maisha, and
  Farid}]{islam2023distinguishing}
Niful Islam, Debopom Sutradhar, Humaira Noor, Jarin~Tasnim Raya,
  Monowara~Tabassum Maisha, and Dewan~Md Farid. 2023.
\newblock Distinguishing human generated text from chatgpt generated text using
  machine learning.
\newblock \emph{arXiv preprint arXiv:2306.01761}.

\bibitem[{Jeblick et~al.(2022)Jeblick, Schachtner, Dexl, Mittermeier,
  St{\"u}ber, Topalis, Weber, Wesp, Sabel, Ricke et~al.}]{jeblick2022chatgpt}
Katharina Jeblick, Balthasar Schachtner, Jakob Dexl, Andreas Mittermeier,
  Anna~Theresa St{\"u}ber, Johanna Topalis, Tobias Weber, Philipp Wesp, Bastian
  Sabel, Jens Ricke, et~al. 2022.
\newblock Chatgpt makes medicine easy to swallow: An exploratory case study on
  simplified radiology reports.
\newblock \emph{arXiv preprint arXiv:2212.14882}.

\bibitem[{Kanda et~al.(2008)Kanda, Miyashita, Osada, Haikawa, and
  Ishiguro}]{Kanda2008}
Takayuki Kanda, Takahiro Miyashita, Taku Osada, Yuji Haikawa, and Hiroshi
  Ishiguro. 2008.
\newblock \href {https://doi.org/10.1109/IROS.2005.1544960} {Analysis of
  humanoid appearances in human-robot interaction.}
\newblock \emph{IEEE Transactions on Robotics}, 24:725--735.

\bibitem[{Kasirzadeh and Gabriel(2023)}]{kasirzadeh2023conversation}
Ardavan Kasirzadeh and Iason Gabriel. 2023.
\newblock In conversation with artificial intelligence: aligning language
  models with human values.
\newblock \emph{Philosophy \& Technology}, 36(2):1--24.

\bibitem[{Kecskes and Zhang(2009)}]{Kecskes2009Activating}
Istvan Kecskes and F.~Zhang. 2009.
\newblock \href {https://doi.org/10.1075/p&c.17.2.06kec} {Activating, seeking,
  and creating common ground: A socio-cognitive approach}.
\newblock \emph{Pragmatics \& Cognition}, 172:331--355.

\bibitem[{Kim et~al.(2023)Kim, Taylor, and Kang}]{Kim2023IsTP}
Zae~Myung Kim, David~E. Taylor, and Dongyeop Kang. 2023.
\newblock "is the pope catholic?" applying chain-of-thought reasoning to
  understanding conversational implicatures.
\newblock \emph{ArXiv}, abs/2305.13826.

\bibitem[{Kong et~al.(2023)Kong, Zhao, Chen, Li, Qin, Sun, and
  Zhou}]{kong2023better}
Aobo Kong, Shiwan Zhao, Hao Chen, Qicheng Li, Yong Qin, Ruiqi Sun, and Xin
  Zhou. 2023.
\newblock \href {http://arxiv.org/abs/2308.07702} {Better zero-shot reasoning
  with role-play prompting}.

\bibitem[{Lai et~al.(2023)Lai, Ngo, Pouran Ben~Veyseh, Man, Dernoncourt, Bui,
  and Nguyen}]{lai2023chatgpt}
Viet Lai, Nghia Ngo, Amir Pouran Ben~Veyseh, Hieu Man, Franck Dernoncourt,
  Trung Bui, and Thien Nguyen. 2023.
\newblock \href {https://doi.org/10.18653/v1/2023.findings-emnlp.878}
  {{C}hat{GPT} beyond {E}nglish: Towards a comprehensive evaluation of large
  language models in multilingual learning}.
\newblock In \emph{Findings of the Association for Computational Linguistics:
  EMNLP 2023}, pages 13171--13189, Singapore. Association for Computational
  Linguistics.

\bibitem[{Leiter et~al.(2023)Leiter, Zhang, Chen, Belouadi, Larionov, Fresen,
  and Eger}]{leiter2023chatgpt}
Christoph Leiter, Ran Zhang, Yanran Chen, Jonas Belouadi, Daniil Larionov,
  Vivian Fresen, and Steffen Eger. 2023.
\newblock Chatgpt: A meta-analysis after 2.5 months.
\newblock \emph{arXiv preprint arXiv:2302.13795}.

\bibitem[{Lu et~al.(2023)Lu, Zhu, Han, Zhao, Mac~Namee, and Tan}]{lu2022makes}
Jinghui Lu, Dongsheng Zhu, Weidong Han, Rui Zhao, Brian Mac~Namee, and Fei Tan.
  2023.
\newblock \href {https://doi.org/10.18653/v1/2023.acl-long.128} {What makes
  pre-trained language models better zero-shot learners?}
\newblock In \emph{Proceedings of the 61st Annual Meeting of the Association
  for Computational Linguistics (Volume 1: Long Papers)}, pages 2288--2303,
  Toronto, Canada. Association for Computational Linguistics.

\bibitem[{Lu et~al.(2024)Lu, Yu, Zhou, and Zhou}]{lu2024large}
Keming Lu, Bowen Yu, Chang Zhou, and Jingren Zhou. 2024.
\newblock \href {http://arxiv.org/abs/2401.12474} {Large language models are
  superpositions of all characters: Attaining arbitrary role-play via
  self-alignment}.

\bibitem[{Mahowald et~al.(2023)Mahowald, Ivanova, Blank, Kanwisher, Tenenbaum,
  and Fedorenko}]{mahowald2023dissociating}
Kyle Mahowald, Anna~A Ivanova, Idan~A Blank, Nancy Kanwisher, Joshua~B
  Tenenbaum, and Evelina Fedorenko. 2023.
\newblock Dissociating language and thought in large language models: a
  cognitive perspective.
\newblock \emph{arXiv preprint arXiv:2301.06627}.

\bibitem[{Marchetti et~al.(2023)Marchetti, Di~Dio, Cangelosi, Manzi, and
  Massaro}]{Marchetti2023Mind}
Antonella Marchetti, Cinzia Di~Dio, Angelo Cangelosi, Federico Manzi, and
  Davide Massaro. 2023.
\newblock \href {https://doi.org/10.3389/frobt.2023.1189525} {Developing
  chatgpt’s theory of mind}.
\newblock \emph{Frontiers in Robotics and AI}, 10.

\bibitem[{Mitrovi{\'c} et~al.(2023)Mitrovi{\'c}, Andreoletti, and
  Ayoub}]{mitrovic2023chatgpt}
Sandra Mitrovi{\'c}, Davide Andreoletti, and Omran Ayoub. 2023.
\newblock Chatgpt or human? detect and explain. explaining decisions of machine
  learning model for detecting short chatgpt-generated text.
\newblock \emph{arXiv preprint arXiv:2301.13852}.

\bibitem[{Mou and Xu(2017)}]{Mou2017TheMI}
Yi~Mou and Kun Xu. 2017.
\newblock The media inequality: Comparing the initial human-human and human-ai
  social interactions.

\bibitem[{Nass and Moon(2000)}]{Nass2000}
Clifford Nass and Youngme Moon. 2000.
\newblock \href {https://doi.org/10.1111/0022-4537.00153} {Machines and
  mindlessness: Social responses to computers}.
\newblock \emph{Journal of Social Issues}, 56:81--103.

\bibitem[{OpenAI(2023)}]{roleplaying}
OpenAI. 2023.
\newblock {Teaching with {A}{I}}.
\newblock \url{https://openai.com/blog/teaching-with-ai}.
\newblock [Online; accessed 17-September-2023].

\bibitem[{Ouyang et~al.(2022)Ouyang, Wu, Jiang, Almeida, Wainwright, Mishkin,
  Zhang, Agarwal, Slama, Ray et~al.}]{ouyang2022training}
Long Ouyang, Jeffrey Wu, Xu~Jiang, Diogo Almeida, Carroll Wainwright, Pamela
  Mishkin, Chong Zhang, Sandhini Agarwal, Katarina Slama, Alex Ray, et~al.
  2022.
\newblock Training language models to follow instructions with human feedback.
\newblock \emph{Advances in Neural Information Processing Systems},
  35:27730--27744.

\bibitem[{Pegoraro et~al.(2023)Pegoraro, Kumari, Fereidooni, and
  Sadeghi}]{pegoraro2023chatgpt}
Alessandro Pegoraro, Kavita Kumari, Hossein Fereidooni, and Ahmad-Reza Sadeghi.
  2023.
\newblock To chatgpt, or not to chatgpt: That is the question!
\newblock \emph{arXiv preprint arXiv:2304.01487}.

\bibitem[{Piantadosi(2023)}]{piantadosia2023modern}
Steven Piantadosi. 2023.
\newblock Modern language models refute chomsky’s approach to language.
\newblock \emph{Lingbuzz Preprint, lingbuzz}, 7180.

\bibitem[{Qin et~al.(2023)Qin, Zhang, Zhang, Chen, Yasunaga, and
  Yang}]{qin2023chatgpt}
Chengwei Qin, Aston Zhang, Zhuosheng Zhang, Jiaao Chen, Michihiro Yasunaga, and
  Diyi Yang. 2023.
\newblock \href {https://doi.org/10.18653/v1/2023.emnlp-main.85} {Is
  {C}hat{GPT} a general-purpose natural language processing task solver?}
\newblock In \emph{Proceedings of the 2023 Conference on Empirical Methods in
  Natural Language Processing}, pages 1339--1384, Singapore. Association for
  Computational Linguistics.

\bibitem[{Qiu et~al.(2023)Qiu, DUAN, and Cai}]{qiu_duan_cai_2023}
Zhuang Qiu, XUFENG DUAN, and Zhenguang~G Cai. 2023.
\newblock \href {https://doi.org/10.31234/osf.io/qtbh9} {Pragmatic implicature
  processing in chatgpt}.

\bibitem[{Radford et~al.(2018)Radford, Narasimhan, Salimans, Sutskever
  et~al.}]{radford2018improving}
Alec Radford, Karthik Narasimhan, Tim Salimans, Ilya Sutskever, et~al. 2018.
\newblock Improving language understanding by generative pre-training.

\bibitem[{Radford et~al.(2019)Radford, Wu, Child, Luan, Amodei, Sutskever
  et~al.}]{radford2019language}
Alec Radford, Jeffrey Wu, Rewon Child, David Luan, Dario Amodei, Ilya
  Sutskever, et~al. 2019.
\newblock Language models are unsupervised multitask learners.
\newblock \emph{OpenAI blog}, 1(8):9.

\bibitem[{Ruis et~al.(2023)Ruis, Khan, Biderman, Hooker, Rocktäschel, and
  Grefenstette}]{ruis2022large}
Laura Ruis, Akbir Khan, Stella Biderman, Sara Hooker, Tim Rocktäschel, and
  Edward Grefenstette. 2023.
\newblock \href {http://arxiv.org/abs/2210.14986} {The goldilocks of pragmatic
  understanding: Fine-tuning strategy matters for implicature resolution by
  llms}.

\bibitem[{Searle(1980)}]{searle1984minds}
John~R Searle. 1980.
\newblock \href {http://cogprints.org/7150/1/10.1.1.83.5248.pdf} {Minds,
  brains, and programs}.
\newblock \emph{Behavioral and Brain Sciences}, 3(3):417--457.

\bibitem[{Shahriar and Hayawi(2023)}]{shahriar2023let}
Sakib Shahriar and Kadhim Hayawi. 2023.
\newblock Let's have a chat! a conversation with chatgpt: Technology,
  applications, and limitations.
\newblock \emph{arXiv preprint arXiv:2302.13817}.

\bibitem[{Shanahan et~al.(2023)Shanahan, McDonell, and
  Reynolds}]{shanahan2023roleplay}
Murray Shanahan, Kyle McDonell, and Laria Reynolds. 2023.
\newblock \href {http://arxiv.org/abs/2305.16367} {Role-play with large
  language models}.

\bibitem[{Susnjak(2022)}]{susnjak2022chatgpt}
Teo Susnjak. 2022.
\newblock Chatgpt: The end of online exam integrity?
\newblock \emph{arXiv preprint arXiv:2212.09292}.

\bibitem[{Sydorenko et~al.(in press)Sydorenko, Dombi, Agrawal, Thorne, Lee, and
  Tao}]{Sydorenko2023}
Tetyana Sydorenko, Judit Dombi, Ameeta Agrawal, Steve Thorne, Jung~In Lee, and
  Yufei Tao. in press.
\newblock Spoken dialogue systems and chatgpt for second language pragmatics
  research.
\newblock In K.~Sadeghi, editor, \emph{The Routledge handbook of technological
  advances and considerations in second language/applied linguistics research}.
  Routledge.
\newblock In press.

\bibitem[{Tabone and De~Winter(2023)}]{tabone2023using}
Wilbert Tabone and Joost De~Winter. 2023.
\newblock Using chatgpt for human--computer interaction research: A primer.
\newblock \emph{Manuscript submitted for publication}.

\bibitem[{Thorp(2023)}]{thorp2023chatgpt}
H~Holden Thorp. 2023.
\newblock Chatgpt is fun, but not an author.
\newblock \emph{Science}, 379(6630):313--313.

\bibitem[{Van~Dis et~al.(2023)Van~Dis, Bollen, Zuidema, van Rooij, and
  Bockting}]{van2023chatgpt}
Eva~AM Van~Dis, Johan Bollen, Willem Zuidema, Robert van Rooij, and Claudi~L
  Bockting. 2023.
\newblock Chatgpt: five priorities for research.
\newblock \emph{Nature}, 614(7947):224--226.

\bibitem[{Wang et~al.(2023)Wang, Hu, Hou, Chen, Zheng, Wang, Yang, Huang, Ye,
  Geng et~al.}]{wang2023robustness}
Jindong Wang, Xixu Hu, Wenxin Hou, Hao Chen, Runkai Zheng, Yidong Wang, Linyi
  Yang, Haojun Huang, Wei Ye, Xiubo Geng, et~al. 2023.
\newblock On the robustness of chatgpt: An adversarial and out-of-distribution
  perspective.
\newblock \emph{arXiv preprint arXiv:2302.12095}.

\bibitem[{Weizenbaum(1966)}]{weizenbaum1966eliza}
Joseph Weizenbaum. 1966.
\newblock Eliza—a computer program for the study of natural language
  communication between man and machine.
\newblock \emph{Communications of the ACM}, 9(1):36--45.

\bibitem[{Wolters et~al.(2009)Wolters, Georgila, Moore, Logie, MacPherson, and
  Watson}]{Wolters2009ReducingWM}
Maria~Klara Wolters, Kallirroi Georgila, Johanna~D. Moore, Robert~H. Logie,
  Sarah~E. MacPherson, and Matthew Watson. 2009.
\newblock Reducing working memory load in spoken dialogue systems.
\newblock \emph{Interact. Comput.}, 21:276--287.

\bibitem[{Zhuo et~al.(2023)Zhuo, Huang, Chen, and Xing}]{zhuo2023exploring}
Terry~Yue Zhuo, Yujin Huang, Chunyang Chen, and Zhenchang Xing. 2023.
\newblock Exploring ai ethics of chatgpt: A diagnostic analysis.
\newblock \emph{arXiv preprint arXiv:2301.12867}.

\end{thebibliography}
